\documentclass[preprint,12pt]{elsarticle}

\usepackage{amssymb}
\usepackage{amsmath}
\usepackage[T1]{fontenc}
\usepackage{graphicx}
\usepackage{array}
\usepackage{tabularx}
\usepackage{booktabs}
\usepackage{multirow}
\usepackage{adjustbox}
\usepackage[table]{xcolor}
\usepackage{colortbl}
\usepackage{caption}
\usepackage{multicol}
\usepackage{arydshln}
\usepackage[colorlinks=true,
            linkcolor=blue,
            anchorcolor=blue,
            citecolor=blue,
            urlcolor=blue]{hyperref}
\newcolumntype{Y}{>{\hspace*{-1.5mm}\centering\arraybackslash}X}

\journal{Medical Image Analysis}

\begin{document}

\begin{frontmatter}
\title{No Modality Left Behind: Adapting to Missing Modalities via Knowledge Distillation for\\ Brain Tumor Segmentation\tnoteref{pubnote}}

\tnotetext[pubnote]{This is the author accepted manuscript. The final published version is available in \textit{Medical Image Analysis}, Volume 112, Article 104108 (2026), doi: \href{https://doi.org/10.1016/j.media.2026.104108}{10.1016/j.media.2026.104108}. In accordance with Elsevier's sharing policy, this arXiv version should be distributed under the CC BY-NC-ND license.}


\author[hdu]{Shenghao Zhu\corref{equal1}}
\author[thu]{Yifei Chen\corref{equal1}}
\author[hdu]{Weihong Chen}
\author[hdu]{Shuo Jiang}
\author[hdu]{Guanyu Zhou}
\author[hdu]{Yuanhan Wang}
\author[hdu]{Feiwei Qin\corref{cor1}}
\author[wcm]{Changmiao Wang}
\author[thu]{Qiyuan Tian\corref{cor1}}

\affiliation[hdu]{organization={Hangzhou Dianzi University},
    city={Hangzhou},
    country={China}}

\affiliation[thu]{organization={Tsinghua University},
            city={Beijing},
            country={China}}

\affiliation[wcm]{organization={Shenzhen Research Institute of Big Data},
            city={Shenzhen},
            country={China}}

\cortext[equal1]{These authors contributed equally to this work.}
\cortext[cor1]{Corresponding authors: Feiwei Qin (qinfeiwei@hdu.edu.cn), Qiyuan Tian (qiyuantian@tsinghua.edu.cn).}

\begin{abstract}
Accurate brain tumor segmentation is essential for preoperative evaluation and personalized treatment. Multi-modal MRI is widely used due to its ability to capture complementary tumor features across different sequences. However, in clinical practice, missing modalities are common, limiting the robustness and generalizability of existing deep learning methods that rely on complete inputs, especially under non-dominant modality combinations. To address this, we propose AdaMM, a multi-modal brain tumor segmentation framework tailored for missing-modality scenarios, centered on knowledge distillation and composed of three synergistic modules. The Graph-guided Adaptive Refinement Module explicitly models semantic associations between generalizable and modality-specific features, enhancing adaptability to modality absence. The Bi-Bottleneck Distillation Module transfers structural and textural knowledge from teacher to student models via global style matching and adversarial feature alignment. The Lesion-Presence-Guided Reliability Module predicts prior probabilities of lesion types through an auxiliary classification task, effectively suppressing false positives under incomplete inputs. Extensive experiments on the Pretreat-MetsToBrain-Masks and BraTS 2018, 2024 datasets demonstrate that AdaMM consistently outperforms existing methods, exhibiting superior segmentation accuracy and robustness, particularly in single-modality and weak-modality configurations. In addition, we conduct a systematic evaluation of six categories of missing-modality strategies, supporting the superiority of knowledge distillation and offering practical guidance for method selection and future research. Our source code is available at \href{https://github.com/Quanato607/AdaMM}{https://github.com/Quanato607/AdaMM}.

\end{abstract}



\end{frontmatter}



\section{Introduction}
With the increasing prevalence of brain tumors, precise segmentation from neuroimaging has become an essential prerequisite for effective diagnosis, individualized therapeutic strategies, and long-term outcome prediction \cite{c1havaei2017brain, c48rasool2025critical}. Magnetic resonance imaging (MRI) offers high-resolution visualization of soft-tissue lesions and captures rich, complementary tumor characteristics across multiple imaging modalities. Specifically, T1-weighted imaging (T1) excels at delineating anatomical structures and vascular edema in subacute stroke, whereas T2-weighted imaging (T2) effectively depicts tissue edema and lesion boundaries \cite{c20verhaert2011direct, c47kolokythas2024t1}. Contrast-enhanced T1-weighted imaging (T1Gd) reveals intratumoral vasculature and blood-brain-barrier disruption via gadolinium enhancement, and fluid-attenuated inversion recovery (FLAIR) comprehensively visualizes abnormal signals associated with stroke and tumor infiltration \cite{c19sarkaria2018blood}. These modalities not only reflect distinct pathological features but also exhibit varying sensitivity to tumor subregions such as the enhancing core, edema (ED), and necrotic areas. The information provided by these modalities is strongly complementary; their combined use furnishes critical evidence for accurate localization, morphological characterization, and fine-grained segmentation of brain tumors. Consequently, multi-modal MRI has become the method of choice for visual analysis and automated segmentation of brain tumors \cite{c30bauer2013survey, c46zhou2024m2gcnet}.

In recent years, advances in automated brain tumor segmentation algorithms have substantially reduced the manual effort required for delineation and improved segmentation accuracy \cite{c41he2025diffusion, c42lu2025fine}. Nevertheless, MRI acquisition in clinical practice is often constrained, and only a subset of modalities is typically available. In contrast, most existing algorithms rely heavily on complete multimodal inputs and exhibit pronounced performance degradation when any modality is missing. Accordingly, a brain-tumor segmentation approach that remains robust to modality absence and is tailored to diverse modality configurations is urgently needed. Traditional generic segmentation models, which lack dedicated modeling for specific modality combinations, frequently display notable instability across different subsets \cite{NEURIPS2024_e95eb520, c15chen2024scunet++}. For example, the T1Gd modality, which has a strong capability for Enhancing Tumor (ET) segmentation, achieves satisfactory results even under single-modality input, wh
ereas the T1 and T2 modalities, which are relatively less effective for ET segmentation, perform poorly in their respective single-modality settings and still yield limited performance when jointly inferred \cite{c17zhang2024tc, c18chen2024toward}. These observations indicate that generic models tend to optimize parameters for modalities with prominent segmentation cues, thus attaining superior performance on dominant modalities while failing to adapt to weaker combinations and consequently impairing overall performance. Hence, generic strategies are constrained under modality absence or non-dominant modality combinations, requiring more targeted, structurally adaptive mechanisms to enhance consistency and robustness in multi-modal settings \cite{c43zhang2025dsca}.

To address this challenge, we propose an Adapter-based Missing Modality (AdaMM) Network comprising a Graph-guided Adaptive Refinement Module (GARM), a Bi-Bottleneck Distillation Module (BBDM), and a Lesion-Presence-Guided Reliability Module (LGRM). In GARM, a lightweight 3D residual module based on an Adapter Bank captures combination-specific features, while a graph convolutional network replaces conventional convolutions to address their weakness in modeling irregular and long-range dependencies, thereby modeling structured semantic consistency across modality completion pathways, an objective that has not been systematically explored in existing missing-modality segmentation methods. In BBDM, a Global Style Matching Module (GSME) and a teacher-student discriminator, coupled with adversarial feature alignment and mean-squared-error loss, achieve precise matching and fusion of bottleneck features between teacher and student models, preserving structural information and textural details between missing-modality and complete-modality pathways and markedly improving robustness and accuracy under modality absence. In LGRM, a lightweight auxiliary classification branch predicts the presence probabilities of lesion types, providing priors for segmentation and distillation and effectively suppressing false positives in hollow regions under missing-modality conditions, while introducing for the first time an explicit lesion presence discrimination branch that injects lesion existence priors into the main segmentation task via a tailored loss formulation.\\

Meanwhile, during preliminary investigations of this study, we found that Zhou \textit{et al.} \cite{c29zhou2024multimodal} systematically categorized existing missing-modality learning methods as imputation-based or imputation-free. Imputation-based approaches compensate for missing modalities by generating data, creating features, or retrieving samples. Data generation methods employ generative adversarial networks (GANs) \cite{c2goodfellow2020generative} or variational autoencoders (VAEs) \cite{c3kingma2013auto} to synthesize missing images conditioned on available modalities; however, the synthetic outputs often lack clinical realism and anatomical fidelity. Feature generation methods predict missing modality features via aggregation mechanisms but heavily depend on existing modalities and fail to recover critical diagnostic cues of the absent modality. Sample retrieval strategies select similar, fully modal cases to supplement missing information, but are limited by the quality of the database and entail high computational costs. Imputation-free methods eschew data synthesis and instead enhance performance through multi-task learning, knowledge distillation, or robustness-oriented strategies. Multi-task learning introduces auxiliary tasks to promote robust shared representations but requires careful weight balancing; knowledge distillation transfers knowledge via teacher-student architectures, boosting accuracy yet risking loss of deep semantic information and involving complex design; robustness enhancement leverages data augmentation or adversarial training to improve tolerance to modality absence, but its efficacy depends on meticulous strategy and parameter selection.

These insights motivated a comprehensive comparative study across the six methodological categories, analyzing brain-tumor segmentation under missing-modality scenarios. Accurate delineation of key tumor regions guides preoperative planning and radiotherapy dosage design, making high performance under modalities absent clinically critical. Experiments on the Pretreat-MetsToBrain-Masks \cite{c44ramakrishnan2024large} and BraTS 2018 \cite{c35menze2014multimodal}, 2024 datasets \cite{c36de20242024}, widely regarded as gold standards, reveal pronounced differences, with knowledge distillation methods indicating favorable performance. By transferring knowledge from complete-modality models, distillation effectively preserves crucial features and maintains high segmentation accuracy despite modality loss. Imputation-based methods can synthesize missing modalities, but the clinical relevance of the generated images remains challenging. In contrast, robustness enhancement and multi-task learning improve adaptability, yet still face limitations in complex missing-modality scenarios. These findings substantiate the advantage of knowledge distillation and provide new perspectives for future research.

The main contributions of this work are as follows:
\begin{enumerate}

  \item We design the BBDM, integrating global style matching with adversarial alignment to achieve precise matching of teacher-student bottleneck features, markedly improving segmentation robustness and accuracy under modality absence. This design ensures that both global structural patterns and fine-grained textural cues are faithfully preserved, even when critical modalities are missing.

  \item We propose the GARM, which explicitly models semantic associations between generalizable and modality-specific features via a graph structure, enabling adaptive enhancement and refined fusion across modality combinations. By leveraging graph-based message passing, the module adaptively emphasizes informative modalities and compensates weaker ones, yielding more balanced and robust representations.

  \item We introduce the LGRM, which guides segmentation by predicting prior probabilities of lesion presence, effectively reducing false segmentation under missing modalities and enhancing cross-modal consistency and reliability. The lesion-aware priors act as a soft constraint during inference, steering the network toward anatomically plausible predictions and improving segmentation reliability.

  \item We conduct a comprehensive evaluation across six methodological categories for missing-modality brain tumor segmentation, encompassing over 15 representative models and 15 modality combinations. This large-scale study supports the consistent superiority of knowledge distillation in maintaining segmentation accuracy under diverse and challenging scenarios, while offering a valuable reference for algorithm selection and future methodological development.

\end{enumerate}

\section{Related Work}
\subsection{Imputation-based Methods}
In multi-modal segmentation tasks, Imputation-based methods explicitly complete missing features or data to recover comprehensive multi-modal semantics. By doing so, a model can maintain, or even substantially enhance, its performance when modalities are absent and can introduce new, more realistic samples into the dataset via generative models such as GANs \cite{c2goodfellow2020generative} or VAEs \cite{c3kingma2013auto}; certain feature-generation strategies adapt well to both homogeneous and heterogeneous data. Nevertheless, these approaches come at a cost: generative models are often complex to train and resource-intensive, and low-quality supplementary data may inject noise into the dataset. When training data are limited or model tuning is inadequate, the model readily overfits the generated data, and the generative model itself may suffer mode collapse \cite{c38karras2020training}, diminishing its ability to faithfully capture the true distribution.

\begin{figure}[htbp]
  \centering
  \includegraphics[width=\textwidth]{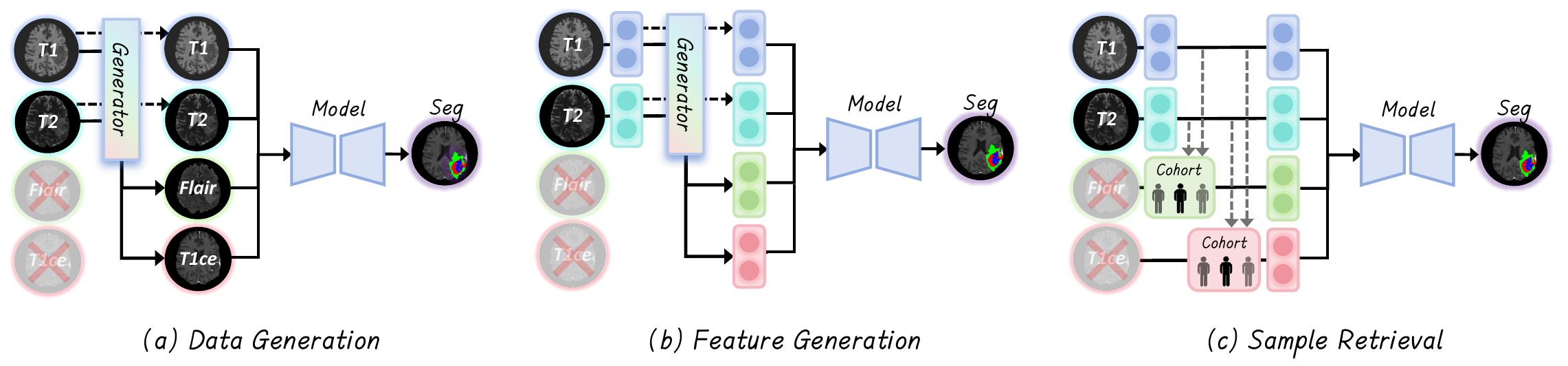}
    \caption{
    (a) \textbf{Data Generation:} an external generator synthesizes absent modalities, creating a full four‑channel input for the segmentation model. 
    (b) \textbf{Feature Generation:} the network learns to hallucinate modality‑specific features internally when inputs are missing. 
    (c) \textbf{Sample Retrieval:} the pipeline retrieves training cases from modality‑matched cohorts to substitute for absent scans before segmentation.}
  \label{fig:imputation_based}
\end{figure}

\subsubsection{Feature Generation}
 These methods first extract shared or contextual features from observed modalities, then synthesize representations in feature space that align with the semantics of the missing modality, and finally perform precise segmentation using the combined features. Representative examples include LCKD by Wang \textit{et al.} \cite{c7wang2023learnable}, which automatically elects the most informative teacher modalities and distills their critical knowledge to student modalities, and ShaSpec proposed by Wang \textit{et al.} \cite{c8wang2023multi}, which supplements missing features through a shared-and-specific dual-branch extractor and a mean-extrapolation strategy. Ting \textit{et al.} \cite{c4ting2023multimodal} further introduced the Multimodal Transformer of Incomplete MRI Data, which explicitly models cross-modal long-range dependencies via self-attention to reconstruct missing sequences at a highly fine granularity. These approaches reduce information loss without synthesizing images, yet they remain insufficiently robust to significant cross-domain distribution shifts and high-dimensional feature noise.

 \subsubsection{Data Generation}
These methods employ generative models (e.g., GANs, VAEs) to first complete the missing modality and then feed the synthesized sequence into the segmentation network. Notable examples include MouseGAN++ by Yu \textit{et al.} \cite{c5yu2022mousegan++}, which leverages a disentangling-contrastive GAN to preserve anatomical structure and contrast, enabling generation of multi-sequence MRI from a single sequence; Hyper-GAE by Yang \textit{et al.} \cite{c6yang2023learning}, which uses a hyper-network and graph-attention mechanism to simultaneously complete all missing sequences in a shared space and optimizes end-to-end with the segmentation network; and SRMNet by Zhang \textit{et al.} \cite{c21li2024deformation}, which introduces deformation-aware reconstruction and key-information mining in the reconstruction branch, substantially enhancing segmentation robustness under complex missing-combination scenarios. Although these methods explicitly complete the inputs, they incur considerable training and inference overhead, and the clinical consistency of the generated modalities remains limited.

 \subsubsection{Sample Retrieval}
Current sample-retrieval methods for handling missing modalities fall into three categories. The first retrieves samples from an external dataset by computing structural or semantic mask similarity (e.g., LRAA \cite{c26fan2024leveraging}), necessitating large data volumes; the second searches within the same batch using mask similarity (e.g., M3Care \cite{c25zhang2022m3care}), which inherently confines retrieval to in-batch samples and struggles with flexible missing-combination scenarios; the third aligns dual modalities of text and images for retrieval (e.g., SMCMSA \cite{c27sun2024similar}), which applies only to classification tasks and is limited to text-image alignment. All three exhibit notably weak robustness to missing modalities typically accommodating only one or two missing modalities and, given the diversity of tumor presentations, classification-oriented retrieval may introduce erroneous information. Consequently, these strategies are largely unsuitable for missing-modality brain-tumor segmentation.

\subsection{Imputation-Free Methods}
In contrast to explicit completion, Imputation-Free methods do not attempt to reconstruct missing modalities; instead, they perform inference directly on incomplete inputs, emphasizing generalization and robustness to any missing-modality combination. Eliminating generation or retrieval markedly reduces computational cost \cite{c40li2025matching} and yields a lightweight inference pipeline, and avoids dependence on a specific completion module, thereby conferring high flexibility across datasets and applications. Robust losses and cooperative mechanisms tailored to missing modalities further endow the model with strong adaptability, enabling solid performance despite incomplete data \cite{c41he2025diffusion}. However, because fine-grained information carried by unobserved modalities cannot be explicitly recovered, the depth of semantic characterization is limited and ultimately hinges on the quality and coverage of the observed data \cite{c39wang2025cross}, meaning that when information is scarce or noisy, results may be incomplete or imprecise.

\begin{figure}[htbp]
  \centering
  \includegraphics[width=\textwidth]{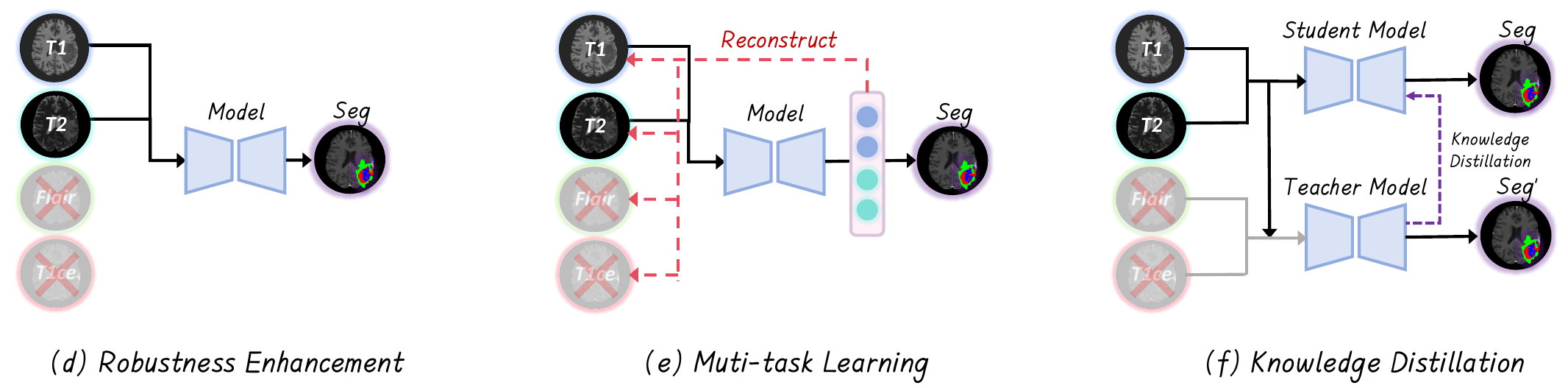}
\caption{
(d) \textbf{Robustness Enhancement:} the network is trained with random modality dropout so it can segment directly from whatever scans are available.  
(e) \textbf{Multi‑task Learning:} an auxiliary decoder reconstructs the absent modalities (red dashed arrows) while the main branch outputs the segmentation mask.  
(f) \textbf{Knowledge Distillation:} a full‑modality teacher guides a partial‑modality student through feature and prediction alignment, boosting accuracy under incomplete inputs.}

  \label{fig:imputation_free}
\end{figure}

\subsubsection{Robust Enhancement}
These methods design robustness-oriented training strategies to strengthen model generalization and adaptability under missing modalities. Ding \textit{et al.} \cite{c11ding2021rfnet} first proposed a Region-aware Fusion Network (RFNet) that adaptively aggregates multi-modal features by exploiting spatial relationships between modalities and tumor regions, significantly enhancing segmentation performance under incomplete modality scenarios.
Zhang \textit{et al.} \cite{c22zhang2022mmformer} subsequently introduced mmFormer, which combines modality-specific encoders with a cross-modal Transformer to capture long-range inter-modal dependencies, enhancing adaptability to arbitrary modality loss. Most recently, Shi \textit{et al.} \cite{c12shi2023mftrans} proposed M$^{2}$FTrans, which incorporates a modality-masking fusion Transformer whose masked self-attention and modality-specific feature re-weighting effectively address different missing-modality scenarios. Although these methods improve robustness, they may still face performance bottlenecks under extreme absence.

\subsubsection{Multi-Task Learning}
These methods optimize multiple related tasks simultaneously, leveraging shared information to enhance model generalization. Chen \textit{et al.} \cite{c10chen2019robust} pioneered a robust multimodal segmentation framework that leverages feature disentanglement, gated fusion, and a reconstruction task, effectively learning modality-invariant representations and significantly enhancing robustness to missing imaging modalities. Jeong \textit{et al.} \cite{c23jeong2022region} later proposed RA-HVED, which uses region-of-interest attention and a joint discriminator for cooperative optimization of segmentation and modality reconstruction. Zhu \textit{et al.} \cite{c9zhu2025xlstm} recently presented XLSTM-HVED, incorporating a visual XLSTM and a cross-perception attention module to reinforce inter-task feature interaction and boost performance when modalities are absent. These methods enhance task cooperation, yet careful tuning of inter-task weights remains necessary.

\subsubsection{Knowledge Distillation}
These methods transfer knowledge from complete-modality models to models operating with missing modalities, thereby substantially enhancing segmentation performance. Wang \textit{et al.} \cite{c13wang2021acn} proposed the Adversarial Collaborative Network (ACN), which uses adversarial learning and inter-modal feature-consistency constraints for robust and efficient teacher-to-student knowledge transfer, improving segmentation accuracy under modality absence. Azad \textit{et al.} \cite{c14ning2021smu} designed Style-Matching UNet (SMUNet), which explicitly transfers discriminative information through content- and style-matching modules, effectively mitigating performance degradation. Zhu \textit{et al.} \cite{c31zhu2025bridging} further proposed the Multi-Scale Transformer Knowledge Distillation Network (MST-KDNet), which fuses a Transformer architecture with multi-scale feature distillation, enabling rich teacher-student feature interaction and significantly improving generalization under missing modalities. Although these methods bolster robustness, more precise capture of inter-modal differential features warrants further exploration.

\section{Method}
We propose a knowledge-distillation-based, multi-modal 3D UNet segmentation framework. The teacher branch performs high-quality segmentation using complete multi-modal inputs, whereas the student branch incorporates lightweight Adapter modules to learn compensation strategies under modality absence and employs the GARM module to enhance feature representation. Both branches share the LGRM module, which leverages lesion-presence probabilities to suppress false positives and applies a voxel-level mean squared error (MSE) distillation constraint on their outputs, allowing the student model to closely approximate teacher performance while remaining computationally and parameter-efficient.

During inference, only the student branch is retained for prediction. The adapter bank selects the corresponding lightweight adapters according to the observed modality combination, GARM performs structured refinement on modality-invariant and combination-aware features near the bottleneck, and LGRM remains active to provide lesion-presence priors for suppressing false positive predictions. In contrast, the teacher branch, GSME, the discriminator, and all distillation-related objectives are used only during training for cross-branch alignment and are completely removed at inference time.\\

\begin{figure}[htbp]
  \centering
  \includegraphics[width=\textwidth]{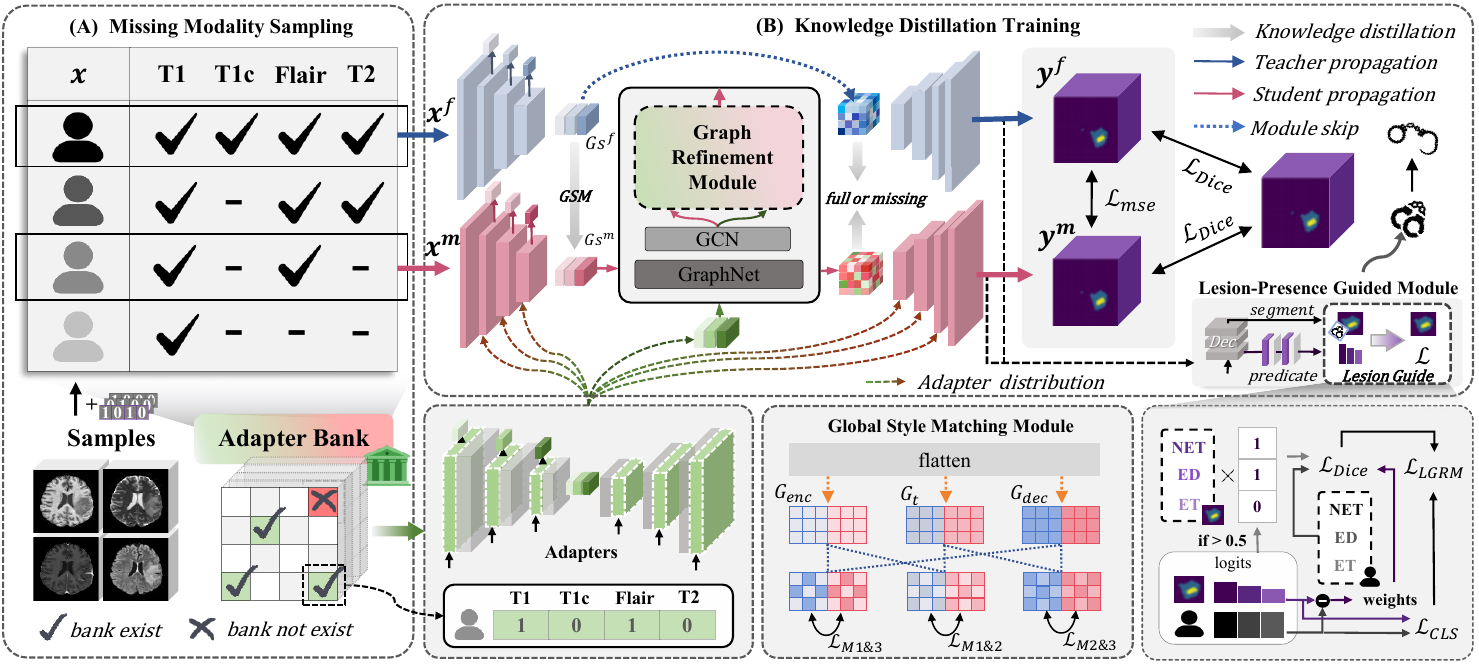}
            \caption{\textbf{Framework overview}. 
    (Stage A) \textbf{Missing‑modality Sampling:} Generates 15 MRI modality combinations and leverages an Adapter Bank to compensate for absent inputs.  
    (Stage B) \textbf{Knowledge‑distillation Training:} Incorporate BBDM, GARM, and LGRM, with GARM applied exclusively to the student branch.}

  \label{fig:framework}
\end{figure}

\subsection{Bi-Bottleneck Distillation Module}
We perform Bi-Bottleneck Distillation Module on the feature before and after the GARM . This knowledge distillation module includes GSME and a teacher-student model discriminator. Due to differences in tissue structure and style features across various MRI modalities, the network decoding process often suffers from information loss in the absence of specific modalities, resulting in a decline in reconstruction quality. To address this issue, we specifically introduce GSME. Concretely, we apply max-pooling to the penultimate encoder features and concatenate them with the bottleneck features; the fused tensor, $\boldsymbol{f}_{\mathrm{enc}}$, is then forwarded to the decoder. The decoder's first convolutional output, $\boldsymbol{f}_{\text{dec}}$, is further processed by reshaping the spatial dimensions $H$, $W$, $D$ of $\boldsymbol{f}_{\text{enc}}, \boldsymbol{f}_{\text{t}}, \boldsymbol{f}_{\text{dec}}$ into two-dimensional tensors $\boldsymbol{G}_{\text{enc}}, \boldsymbol{G}_{\text{t}}, \boldsymbol{G}_{\text{dec}}$. These tensors are then subjected to a subsequent feature fusion operation:
\begin{gather}
    \boldsymbol{M}_{1} = \boldsymbol{G}_{\text{enc}} \boldsymbol{G}_{\text{dec}}^T, \quad
    \boldsymbol{M}_{2} = \boldsymbol{G}_{\text{enc}} \boldsymbol{G}_{\text{t}}^T, \quad
    \boldsymbol{M}_{3} = \boldsymbol{G}_{\text{dec}} \boldsymbol{G}_{\text{t}}^T.
\end{gather}
After obtaining $\boldsymbol{M} \in \{\boldsymbol{M}_1, \boldsymbol{M}_2, \boldsymbol{M}_3\}$, we enforce a MSE distillation loss:
\begin{align}
    \mathcal{L}_{\text{GSME}} &= \theta / (4n^2) \;
    \sum_{i=1}^{3} \sum_{j=1}^{n} \Big( \boldsymbol{M}_{i,j}^{f} - \boldsymbol{M}_{i,j}^{m} \Big)^2,
\end{align}
where superscripts $f$ and $m$ denote teacher and student features, respectively.Once the enhanced features exit the student’s GARM module, we pair them with the corresponding bottleneck features from the teacher and feed both into a discriminator~$D$ to compute the adversarial loss:
\begin{equation}
    \mathcal{L}_{\text{adv}} = \log(1 - D(\boldsymbol{f}_{\text{GARM}}^f)) + \log(D(\boldsymbol{f}_{\text{GARM}}^m)).
\end{equation}
The overall distillation loss for the BBDM is therefore defined as:
\begin{equation}
\mathcal{L}_{\text{BBDM}}
= \lambda\,\mathcal{L}_{\text{adv}}
+ \theta\,\mathcal{L}_{\text{GSME}},
\end{equation}
where $\lambda$ and $\theta$ are balancing hyper-parameters. Joint optimisation of these losses transfers high-level structural knowledge and low-level texture cues from the teacher’s encoder and bottleneck to the student model.  
Consequently, the student preserves cross-modal path consistency and alignment under missing-modality conditions, significantly enhancing overall reconstruction robustness and accuracy.

The proposed distillation strategy does not strongly depend on the teacher model’s predictive quality, as supervision is deliberately restricted to bottleneck-level representations to provide high-level semantic and structural guidance rather than enforcing layer-wise imitation. The student adopts an independent encoder-decoder architecture, where feature completion is primarily driven by modality-specific adapters and graph-guided reasoning across different missing-modality combinations. Adversarial and GSME losses are applied only to enhanced features before decoding, while semantic enhancement and reconstruction are optimized within the student model itself, without direct layer-wise constraints from the teacher. As a result, the teacher mainly stabilizes training instead of defining the student’s performance upper bound, allowing the student to correct potential biases through their own encoding-decoding pathway and structured reasoning, and to learn more robust and generalizable representations under missing-modality conditions.

\begin{figure}[htbp]
  \centering
  \includegraphics[width=\textwidth]{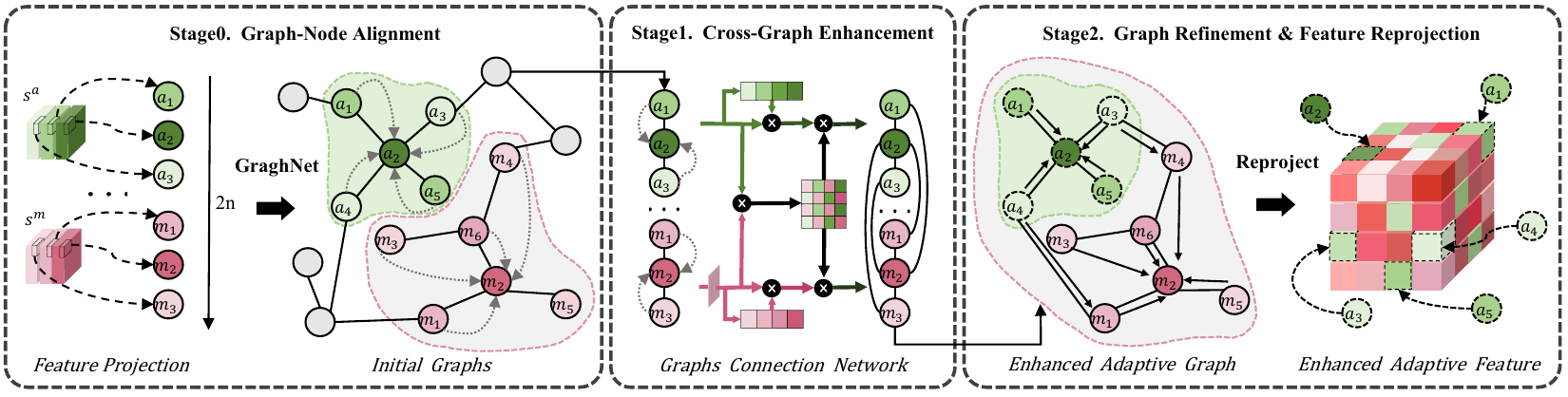}
    \caption{\textbf{Graph-guided Adaptive Refinement Module}. 
    (\textbf{Stage 0}) Graph-node Alignment, which establishes consistent correspondence between nodes across modality-specific and shared graphs. (\textbf{Stage 1}) Cross‑graph Enhancement, which facilitates mutual information exchange to enrich node representations. (\textbf{Stage 2}) Graph Refinement \& Feature Reprojection, which progressively refines node embeddings and projects them back to the feature space for downstream processing.}

  \label{fig:GARM}
\end{figure}

\subsection{Graph-guided Adaptive Refinement Module}
We have innovatively designed the GARM module to explicitly model the semantic association between Generalization Features (GF) and Combination-aware Features (CF), thereby enhancing feature representation and improving knowledge distillation performance. GARM utilizes a graph neural network structure \cite{c28li2024multi} as a bridge to explicitly connect GF and CF, enabling collaborative reasoning and ultimately projecting enhanced semantic information back into the dense feature space to refine segmentation predictions. In the specific implementation, we first constructed a training process that simulates missing modalities, generating multiple modality combinations from full-modality inputs using 15 random masking strategies, with missing modalities treated as zero-filled. The processed input \( x \in \mathbb{R}^{H \times W \times D \times C} \) is first processed through 3D convolutions to obtain initial features, then progressively reduced in spatial dimension through three layers of stride=2 downsampling convolutions. The GARM module consists of the Adapter Bank and three stages: Graph-Node Alignment, Cross-Graph Enhancement, and Graph Refinement \& Feature Reprojection.

{
\makeatletter
\renewcommand\thesubsubsection{%
  \thesubsection.\@arabic\c@subsubsection}
\makeatother
\subsubsection{Adapter Bank}
}

The number of modality-specific adapters is inherently bounded in the target application setting. In a typical three-dimensional MRI brain tumor segmentation task, the number of available modalities is usually no more than four. As a result, the total number of possible missing-modality combinations is at most $2^4-1$, which equals 15. This scale is fixed and does not increase with dataset size or the number of training samples, thereby ensuring that the adapter library remains controllable in practical use. In addition, each adapter in the Adapter Bank is constructed from lightweight three-dimensional residual modules, whose parameter count is negligible compared to that of the backbone network. Even when dedicated adapters are instantiated for all modality combinations, the resulting increase in overall model parameters and computational overhead remains very limited and does not impose a meaningful burden on training or inference efficiency.

More importantly, maintaining independent adapters for different modality combinations enables the model to learn targeted compensation strategies tailored to specific missing-modality patterns, without relying on a highly coupled and higher-capacity shared module to handle all cases simultaneously. This form of combination-level adaptation is particularly advantageous in missing-modality scenarios, as it helps mitigate instability induced by feature distribution discrepancies and further improves both robustness and the achievable performance ceiling under complex missing conditions.

\subsubsection{Graph-node Alignment}
Given the CF voxel features $\boldsymbol{F}_{\mathrm{c}}\in\mathbb{R}^{(8H\times8W\times8D\times32C)}$ from the adapter and the GF features $\boldsymbol{F}_{\mathrm{g}}\in\mathbb{R}^{(8H\times8W\times8D\times32C)}$ obtained via 3D convolution, we instantiate two \texttt{GraphNet3D} modules (for CF and GF) to parameterize $K$ learnable anchors and channel-wise bandwidths and to project dense voxels into graph form. Each instance maintains an anchor matrix as the node set and supports HDF5-based initialization of anchors. The structures are:
\begin{equation}
\left\{
\begin{aligned}
\mathcal{G}_c &= (\boldsymbol{G}_c,\;\boldsymbol{P}_c),\quad
\boldsymbol{G}_c\in\mathbb{R}^{K\times 32C},\quad
\boldsymbol{P}_c\in\mathbb{R}^{K\times N},\\
\mathcal{G}_g &= (\boldsymbol{G}_g,\;\boldsymbol{P}_g),\quad
\boldsymbol{G}_g\in\mathbb{R}^{K\times 32C},\quad
\boldsymbol{P}_g\in\mathbb{R}^{K\times N},
\end{aligned}
\right.
\end{equation}
where $\boldsymbol{G}_\star$ stacks the learnable anchors for branch $\star\in\{c,g\}$ and serves as node features, and $\boldsymbol{P}_\star$ is the soft-assignment matrix with $N{=}8H\cdot8W\cdot8D$. Additionally, \texttt{GraphNet3D} is a learnable projection module that maintains $K$ trainable anchor vectors (one per graph node) and corresponding channel-wise bandwidth parameters. Given a voxel feature map, it flattens the spatial dimensions, computes channel-wise Gaussian distances between each voxel and each anchor, and applies a softmax over nodes to produce a voxel-to-node soft-assignment matrix $\boldsymbol{P}$. This assignment, together with the anchor matrix $\boldsymbol{G}$, defines the graph representation of the input features.

\begin{figure}[!t]
  \centering
  \includegraphics[width=\textwidth]{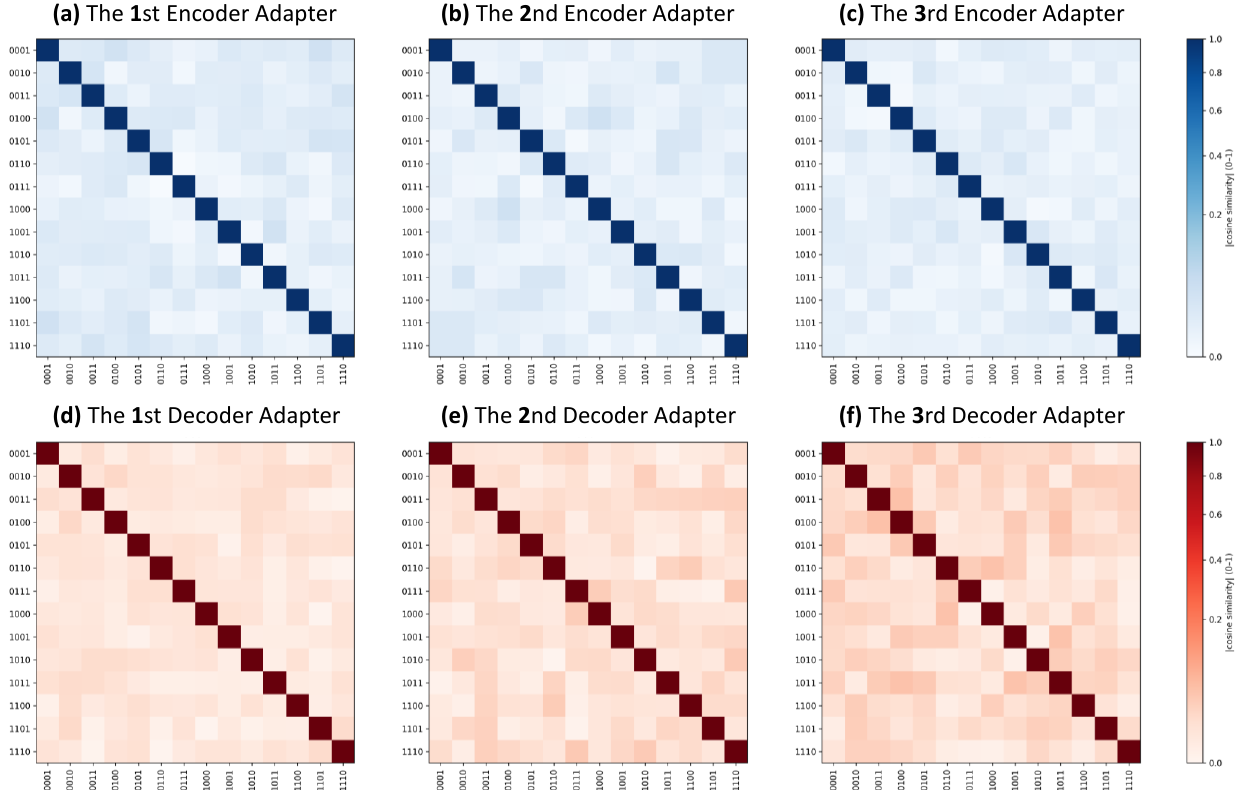}
    \caption{
    \textbf{Cosine similarity heatmaps of adapter parameters.} 
    Each four-digit code denotes available modalities in the order [T1, T1Gd, T2, FLAIR] (1=present, 0=absent). 
    \textbf{(a-c)} \textbf{1}st-\textbf{3}rd encoder adapters, \textbf{(d-f)} \textbf{1}st-\textbf{3}rd decoder adapters. 
    Deeper adapters show higher inter-configuration similarity, indicating convergence toward consistent full-modality feature approximation, 
    while retaining clear modality-combination specificity.
    }

  \label{fig:GARM_sim}
\end{figure}

For any voxel feature $f_i$, the soft assignment to the $k$-th node is computed by a channel-wise Gaussian kernel with learnable bandwidth:
\begin{equation}
\boldsymbol{P}_{k,i}
= \frac{\exp\!\Bigl(-\bigl\|\,(f_i-\mu_k)\odot\boldsymbol{\sigma}_k^{-1}\,\bigr\|_2^2/2\Bigr)}
       {\sum_{j=1}^{K}\exp\!\Bigl(-\bigl\|\,(f_i-\mu_j)\odot\boldsymbol{\sigma}_j^{-1}\,\bigr\|_2^2/2\Bigr)},
\end{equation}
where $f_i,\mu_k\in\mathbb{R}^{32C}$ denote the voxel feature and the $k$-th learnable anchor, $\boldsymbol{\sigma}_k\in\mathbb{R}^{32C}$ is a learnable channel-wise bandwidth vector, and $\odot$ denotes element-wise scaling by $\boldsymbol{\sigma}_k^{-1}$. By mapping high-dimensional voxels to a compact $K$-node space with learnable anchors and adaptive bandwidths, Graph-node Alignment forms $\boldsymbol{P}$ as a locality-preserving correspondence from voxels to nodes and provides a graph-based abstraction that prepares GF and CF for subsequent cross-graph reasoning with reduced voxel-level redundancy.

\subsubsection{Cross-Graph Enhancement}

To impose explicit one-to-one correspondence prior to holistic graph reasoning, we perform an efficient alignment between the GF and CF graphs. Specifically, a channel-wise linear transformation is applied to the CF graph to harmonize its feature scale with the GF graph:
\begin{equation}
\boldsymbol{\tilde{G}}_{c}
= \boldsymbol{W}_{c}\,\boldsymbol{G}_{c},
\quad
 \boldsymbol{W}_{c}\in\mathbb{R}^{32C\times32 C}.
\end{equation}

We then establish a comprehensive and fully connected similarity matrix between the GF and CF nodes, and use $\boldsymbol{E}$ as the neighborhood to propagate information along GF and CF edges via a graph convolution operation:
\begin{equation}
\begin{gathered}
\boldsymbol{E} = \boldsymbol{G}_{g}^{\top}\,\boldsymbol{\tilde{G}}_{c} \in \mathbb{R}^{K\times K},\\
\boldsymbol{G'}_{g} = \operatorname{GCN}(\boldsymbol{G}_{g}, \boldsymbol{E}), 
\quad
\boldsymbol{G'}_{c} = \operatorname{GCN}(\boldsymbol{\tilde{G}}_{c}, \boldsymbol{E}),
\end{gathered}
\end{equation}
where \texttt{GCN} refers to a graph convolutional layer that first applies a learnable linear transformation to each node’s feature vector, then normalizes the adjacency matrix $\boldsymbol{E}$ with a softmax along the neighbor dimension, and finally aggregates transformed features from connected nodes according to these normalized weights. This operation enables information exchange between nodes while respecting the similarity structure encoded in $\boldsymbol{E}$. 

By enforcing a one-to-one correspondence between GF and CF nodes through scale-aligned similarity and cross-graph GCN propagation, this block harmonizes the two feature spaces, strengthens correspondence semantics, and improves the stability of distilled representations as well as the accuracy of downstream segmentation under diverse modality combinations.

\subsubsection{Graph Refinement and Feature Reprojection}

To capture complementary semantics from GF and CF, we compute an inter-node cosine similarity matrix to construct a cross-graph adjacency matrix $\boldsymbol{A}$ by thresholding the similarity matrix $\boldsymbol{S}$ with a threshold value $\tau = 0.8$:
\begin{equation}
\boldsymbol{S} = \bigl[\cos(\boldsymbol{v}_{p}^{g},\,\boldsymbol{v}_{q}^{c})\bigr]_{p,q=1}^{K},
\quad \boldsymbol{v}_{p}^{c}\in \boldsymbol{G'}_{c},\ \boldsymbol{v}_{g}^{c}\in \boldsymbol{G'}_{g},
\end{equation}
\begin{equation}
\boldsymbol{A}_{p,q}
= 
\begin{cases}
1, & \boldsymbol{S}_{p,q} > \tau,\\
0, & \text{otherwise}.
\end{cases}, \quad
\boldsymbol{A}' = (\boldsymbol{A} + \boldsymbol{A}^{\top}) / 2,
\end{equation}
where $\boldsymbol{v}$ denotes the node feature vector in $\boldsymbol{G'}$.Then Graph Attention Networks (GAT) \cite{c37velickovic2018graph} are utilized for performing cross-graph aggregation and we build a unified adjacency to facilitate deeper cross-graph interactions:
\begin{equation}
\begin{gathered}
\boldsymbol{G''}_{g} = \operatorname{GAT}(\boldsymbol{G'}_{g}, \boldsymbol{A}), 
\quad
\boldsymbol{G''}_{c} = \operatorname{GAT}(\boldsymbol{G'}_{c}, \boldsymbol{A}),\\[6pt]
\boldsymbol{A}'' =
\begin{bmatrix}
\boldsymbol{A} & \boldsymbol{A}' \\[6pt]
\boldsymbol{A}' & \boldsymbol{A}^\top
\end{bmatrix}
\in\mathbb{R}^{2K\times 2K}, \quad
\boldsymbol{A}''' 
= (\boldsymbol{A}'' + {\boldsymbol{A}''}^{\top}) / 2.
\end{gathered}
\end{equation}
Joint aggregation is performed on the concatenated node features:
\begin{equation}
\boldsymbol{G}_{g,c} =
[\boldsymbol{G'''}_{g} ;\boldsymbol{G'''}_{c} ]= \operatorname{GAT}([\boldsymbol{G''}_{g} ;\boldsymbol{G''}_{c} ], \boldsymbol{A}''').
\end{equation}

After the two-stage cross-graph aggregation, the updated node features encode both intra- and inter-graph semantics. These features are projected back to the dense voxel space via the soft-assignment matrices $\boldsymbol{P}_{c}$ and $\boldsymbol{P}_{g}$, with each node feature distributed to all voxels according to normalized weights $\boldsymbol{w}_{c}$ and $\boldsymbol{w}_{g}$ derived from $\boldsymbol{P}_{c}$ and $\boldsymbol{P}_{g}$:
\begin{equation}
\begin{gathered}
[\boldsymbol{w}_c; \boldsymbol{w}_g] = \operatorname{softmax}([\boldsymbol{P}_c;\boldsymbol{P}_g]), \quad
\boldsymbol{P}_{g,c} = \boldsymbol{P}_{g} \cdot \boldsymbol{w}_g + \boldsymbol{P}_{c} \cdot \boldsymbol{w}_c,
\\
\mathcal{G}_{g,c} = \bigl(\boldsymbol{G'''}_c,\,\boldsymbol{P}_{g,c}\bigr),
\quad
\boldsymbol{G'''}_{c}\in\mathbb{R}^{K\times 32C},
\quad
\boldsymbol{P}_{g,c}\in\mathbb{R}^{K\times N}.
\end{gathered}
\end{equation}
The projected features are further refined by $1\times1\times1$ convolutional layers and fused with the original features via residual addition:
\begin{equation}
\boldsymbol{F} = \boldsymbol{G'''}_{c}^\top\boldsymbol{P}_{g,c}, \quad
\boldsymbol{F'} = \operatorname{Conv}_2(\operatorname{Conv}_1\boldsymbol{F} + \operatorname{Conv}_0(\boldsymbol{F}_g + \boldsymbol{F}_c)).
\end{equation}
Finally, a $1\times1\times1$ convolution is employed to produce the final segmentation prediction. By explicitly modeling the fine-grained semantic interplay between GF and CF through a graph-based mechanism, GARM effectively integrates generalizable and combination-specific cues, thereby enriching the bottleneck feature representation and markedly enhancing the model’s adaptability and expressiveness in multimodal scenarios. 

\subsection{Lesion-Presence Guided Reliability Module}
In the multimodal tumor segmentation task, the probability of the presence of different lesion types (Non-Enhancing Tumor Core (NET), ED, ET in BraTS 2018  \cite{c35menze2014multimodal} or Non-Enhancing Tumor Core (NETC), Surrounding Non-enhancing FLAIR Hyperintensity (SNFH), ET in BraTS 2024  \cite{c36de20242024}) can provide an important prior for the subsequent segmentation with distillation loss, and often the model misclassifies the tumor. Taking this BraTS 2018 as an example, we embed a lightweight auxiliary classification branch at the end of the decoder, uniformly named LGRM, which introduces only two layers of convolution and a single global average pooling layer, with minimal additional overhead, yet significantly improves Dice and cross-modal consistency. We utilize the $\boldsymbol{u}_2$ feature from the decoder's quadratic upsampling to input the LGRM and obtain the probability vectors. The LGRM consists of two layers of $3\times3\times3$ convolution, a BatchNorm, a Global Average Pooling (GAP), and Sigmoid activation:
\begin{equation}
\begin{gathered}
\boldsymbol{z} = \phi\!\bigl(\operatorname{BN}\!\bigl(\operatorname{Conv}_{3\times3\times3}(\boldsymbol{u}_2)\bigr)\bigr), \quad
\boldsymbol{q} = \operatorname{Conv}_{1\times1\times1}(\boldsymbol{z}) \in \mathbb{R}^{3\times D\times H\times W},\\[6pt]
\hat{\boldsymbol{p}} = \sigma\!\bigl(\operatorname{GAP}(\boldsymbol{q})\bigr)
        = \bigl[\hat{p}_{\mathrm{NET}},\,\hat{p}_{\mathrm{ED}},\,\hat{p}_{\mathrm{ET}}\bigr]^{\top}
        \in (0,1)^{3},
\end{gathered}
\end{equation}
where $\phi(\cdot)$ denotes the relu activation, $\sigma(\cdot)$ the sigmoid activation, and $\hat{\boldsymbol{p}}$ the presence probabilities of the three lesion types.  
During inference, to suppress false positives, the three-channel segmentation probability map $\boldsymbol{S} \in [0,1]^{B\times 3 \times D \times H \times W}$ is multiplied by the existence mask $\boldsymbol{g}$:
\begin{equation}
\begin{gathered}
\boldsymbol{g}_{b,i}
=
\begin{cases}
1, & \hat{\boldsymbol{p}}_{b,i} > 0.5,\\
0, & \text{otherwise},
\end{cases}
\quad b=1,\dots,B,\; i\in\{\mathrm{NET},\mathrm{ED},\mathrm{ET}\},\\[6pt]
\boldsymbol{S}^{\ast} = \boldsymbol{g} \otimes \boldsymbol{S},
\end{gathered}
\end{equation}
where $\otimes$ indicates broadcast multiplication along the channel dimension and $\boldsymbol{S}^{\ast}$ is the final segmentation output.  
When $\hat{p}^{i}\le0.5$, the probability volume for the $i$-th class is zeroed out, effectively suppressing false positives in empty regions.
Meanwhile, the vector $\hat{\boldsymbol{p}}$ is used to derive a probability-weighted segmentation loss and a binary cross-entropy (BCE) classification loss.  
Let $\boldsymbol{y}^{\mathrm{cls}}\in\{0,1\}^{B\times3}$ be the hard labels for the three lesion types, obtained by max pooling the ground-truth segmentation masks.

For the complete-modality (full) and missing-modality (miss) branches, the weighting factors are defined as:
\begin{equation}
\boldsymbol{w}_{\text{full}} = \lvert \hat{\boldsymbol{p}}_{\text{full}} - \boldsymbol{y}^{\text{cls}} \rvert + 1,
\quad
\boldsymbol{w}_{\text{miss}} = \lvert \hat{\boldsymbol{p}}_{\text{miss}} - \boldsymbol{y}^{\text{cls}} \rvert + 1.
\end{equation}
The purpose of these weights is intuitive: if the network predicts the presence of a lesion with high confidence ($\hat{\boldsymbol{p}}\approx1$) but the label is $0$, the term $\lvert\hat{\boldsymbol{p}}-\boldsymbol{y}\rvert$ becomes large, the weight increases, and the false-positive error is penalised; conversely, if the network fails to predict a lesion that is actually present ($\hat{\boldsymbol{p}}\approx0$, $\boldsymbol{y}=1$), the weight likewise increases to penalise the false negative.  
The weighted Dice losses and binary cross-entropy (BCE) classification losses for the three lesion types are thus:
\begin{equation}
\begin{gathered}
\mathcal{L}_{\mathrm{Dice}}^{\mathrm{full}}
= \sum_{i\in\{\mathrm{NET},\mathrm{ED},\mathrm{ET}\}}
  \boldsymbol{w}_{\mathrm{full},i} / B \;
  \sum_{b=1}^{B}
  \operatorname{Dice}\!\bigl(\boldsymbol{S}^{(b)}_{i}, \boldsymbol{Y}^{(b)}_{i}\bigr),\\[6pt]
\mathcal{L}_{\mathrm{Dice}}^{\mathrm{miss}}
= \sum_{i\in\{\mathrm{NET},\mathrm{ED},\mathrm{ET}\}}
  \boldsymbol{w}_{\mathrm{miss},i} / B \;
  \sum_{b=1}^{B}
  \operatorname{Dice}\!\bigl(\boldsymbol{S}^{(b)}_{i,\mathrm{miss}}, \boldsymbol{Y}^{(b)}_{i}\bigr),\\[6pt]
\mathcal{L}_{\mathrm{CLS}}^{\mathrm{full}}
= \operatorname{BCE}\!\bigl(\hat{\boldsymbol{p}}_{\mathrm{full}}, \boldsymbol{y}^{\mathrm{cls}}\bigr), \quad
\mathcal{L}_{\mathrm{CLS}}^{\mathrm{miss}}
= \operatorname{BCE}\!\bigl(\hat{\boldsymbol{p}}_{\mathrm{miss}}, \boldsymbol{y}^{\mathrm{cls}}\bigr).
\end{gathered}
\end{equation}
Therefore, the overall LGRM loss is formulated as the sum of the probability-weighted Dice terms and the BCE classification terms for both the full-and missing-modality branches:
\begin{equation}
\mathcal{L}_{\mathrm{LGRM}}
= \mathcal{L}_{\mathrm{Dice}}^{\mathrm{full}}
+ \mathcal{L}_{\mathrm{Dice}}^{\mathrm{miss}}
+ \mathcal{L}_{\mathrm{CLS}}^{\mathrm{full}}
  + \mathcal{L}_{\mathrm{CLS}}^{\mathrm{miss}}.
\end{equation}

\begin{figure}[htbp]
  \centering
  \includegraphics[width=\textwidth]{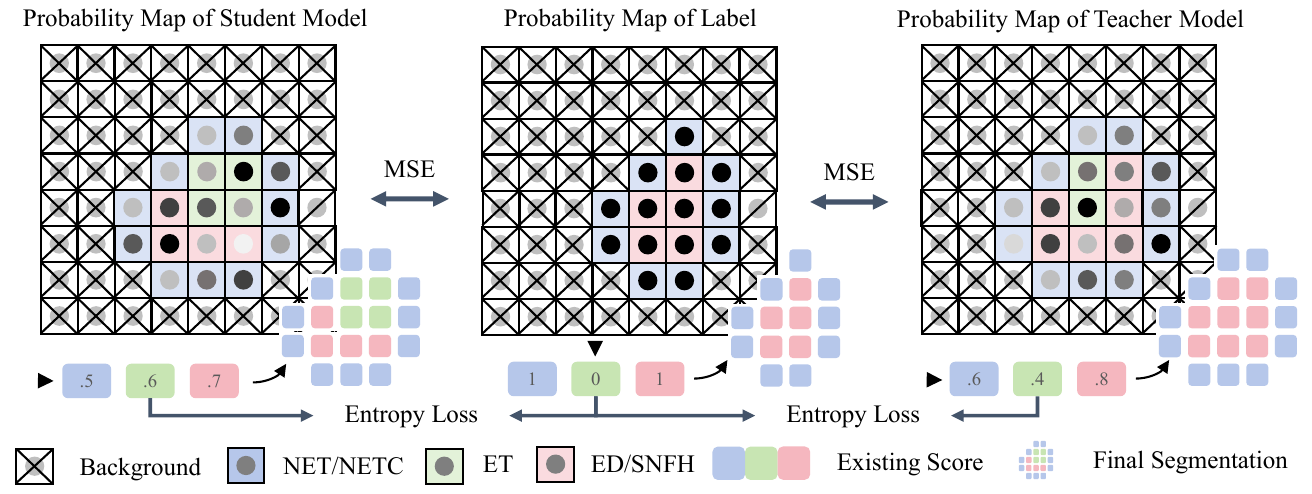}
\caption{\textbf{Lesion‑Presence‑Guided Reliability Module.} 
For each slice, voxel‑level probability maps from the student (left) and teacher (right) are aligned to the ground‑truth label map (centre) with a voxel‑wise mean‑squared‑error.  
Concurrently, the three lesion classes (NET, ET, ED) are collapsed into existence scores; these scores are matched to the binary presence vector of the label via an entropy loss.  }

  \label{fig:lgrm}
\end{figure}

\subsection{Joint Objective}
During the training process, we jointly optimize three types of loss terms to enhance the model's performance, constituting the final total loss function:
\begin{equation}
\mathcal{L}_{\mathrm{joint}}
=\mathcal{L}_{\mathrm{mse}}+
\mathcal{L}_{\mathrm{BBDM}}
+\mathcal{L}_{\mathrm{LGRM}}.
\end{equation}
Among them, $\mathcal{L}_{\mathrm{mse}}$ is the teacher-student predictive distillation loss at the voxel level, which guides the student model to quickly fit the predictive distribution of the teacher model, accelerates convergence and enhances the boundary consistency by applying the MSE to the segmentation probability map; $\mathcal{L}_{\mathrm{BBDM}}$ consists of the MSE of bottleneck features in the GSME module together with the discriminator confrontation loss, which performs the in-depth alignment of structural hierarchies and texture distributions, respectively, to to improve the reconstruction quality and detail restoration ability of the student model in the absence of modality; $\mathcal{L}_{\mathrm{LGRM}}$ introduces the binary cross-entropy loss of classification branch and the probability-weighted Dice loss of segmentation branch based on the existence probability of the lesion in the output of the LGRM module, which explicitly suppresses the false-positives and false-negatives, and especially strengthens the modeling of the boundary ambiguity and the uncertainty region.

\section{Results}
\subsection{Datasets}
\subsubsection{BraTS 2024 \cite{c36de20242024}}
This dataset comprises multimodal MRI scans from 1,350 patients, each case including T1, T2, FLAIR, and T1Gd. All images have been resampled to an isotropic resolution of 1 mm³, registered to a standard anatomical template, and pre-processed with skull stripping and related steps. Experienced radiologists manually annotated each case; the segmentation targets encompass three regions: Whole Tumor (WT), Tumor Core (TC), and ET. We split the dataset 80 \% / 20 \% into training and testing sets to evaluate the model’s representational capacity and scalability.

\subsubsection{BraTS 2018 \cite{c35menze2014multimodal}}
This dataset shares the same four MRI modalities and three annotation regions as BraTS 2024 but contains only 285 cases, making it suitable for testing a model’s generalization and resistance to overfitting under limited data. We apply the same 80 \% / 20 \% training-testing split.

\subsubsection{Pretreat-MetsToBrain-Masks Dataset~\cite{c44ramakrishnan2024large}}

This dataset comprises 200 cases of 3D brain metastasis MRI scans. All segmentation annotations were meticulously delineated by expert radiologists and preprocessed using a standardized protocol. The modality configuration and label taxonomy follow those of BraTS 2018. The dataset exhibits substantial heterogeneity, ranging from sub-centimeter metastases to necrotic lesions with complex boundaries, which poses increased challenges for detection and segmentation. An identical 80 \% / 20 \% training-testing split is adopted.


\subsection{Experiments Detail}
All experiments were conducted on a computing server equipped with NVIDIA Tesla A800 GPUs, and the deep‑learning framework was PyTorch 2.4.1. The batch size was fixed at 1. The model was trained for 400 epochs on the BraTS 2024 dataset and for 1,000 epochs on the BraTS 2018 dataset and Pretreat-MetsToBrain-Masks dataset to exploit its limited samples better.
Model parameters were updated with the Adam optimizer; the initial learning rate was $1\times 10^{-4}$ and decayed to $1\times 10^{-5}$ to ensure training stability and efficient convergence.
For data preprocessing, all MRI volumes were center‑cropped and resampled to an input size of $160\times 192\times 168$ voxels, followed by intensity normalization to promote training stability. During training, we applied several data‑augmentation strategies, including random flipping, rotation, and random cropping to improve generalization. During testing, the images were likewise center‑cropped to the same dimensions to maintain consistency with the training phase. 

We do not adopt decoder weight sharing between the teacher and student models, as this is a deliberate architectural design choice motivated by considerations of stability and generalization. Sharing the decoder would introduce unnecessary parameter coupling during training, causing the optimization process to be biased toward the full-modality distribution and consequently weakening the student model’s ability to perform targeted modeling of missing-modality features. In contrast, employing independent decoders allows the student model to focus on reconstructing semantic representations that have been completed based on the encoder outputs and graph-guided reasoning. When combined with the modality-specific adapter mechanism, this design enables the student model to learn more flexible and stable decoding strategies tailored to different missing-modality combinations. Without parameter sharing, collaboration between the teacher and student is achieved through explicit feature alignment mechanisms. These include style-based feature alignment and discriminator-based constraints applied at intermediate feature levels, as well as graph-guided reasoning that further reduces the distributional discrepancy between missing-modality features and full-modality features. Together, these mechanisms provide a consistent and reliable semantic foundation for subsequent decoding, enabling effective and stable teacher-student collaborative learning without decoder weight sharing.

\begin{figure}[!t]
  \centering
  \includegraphics[width=\textwidth]{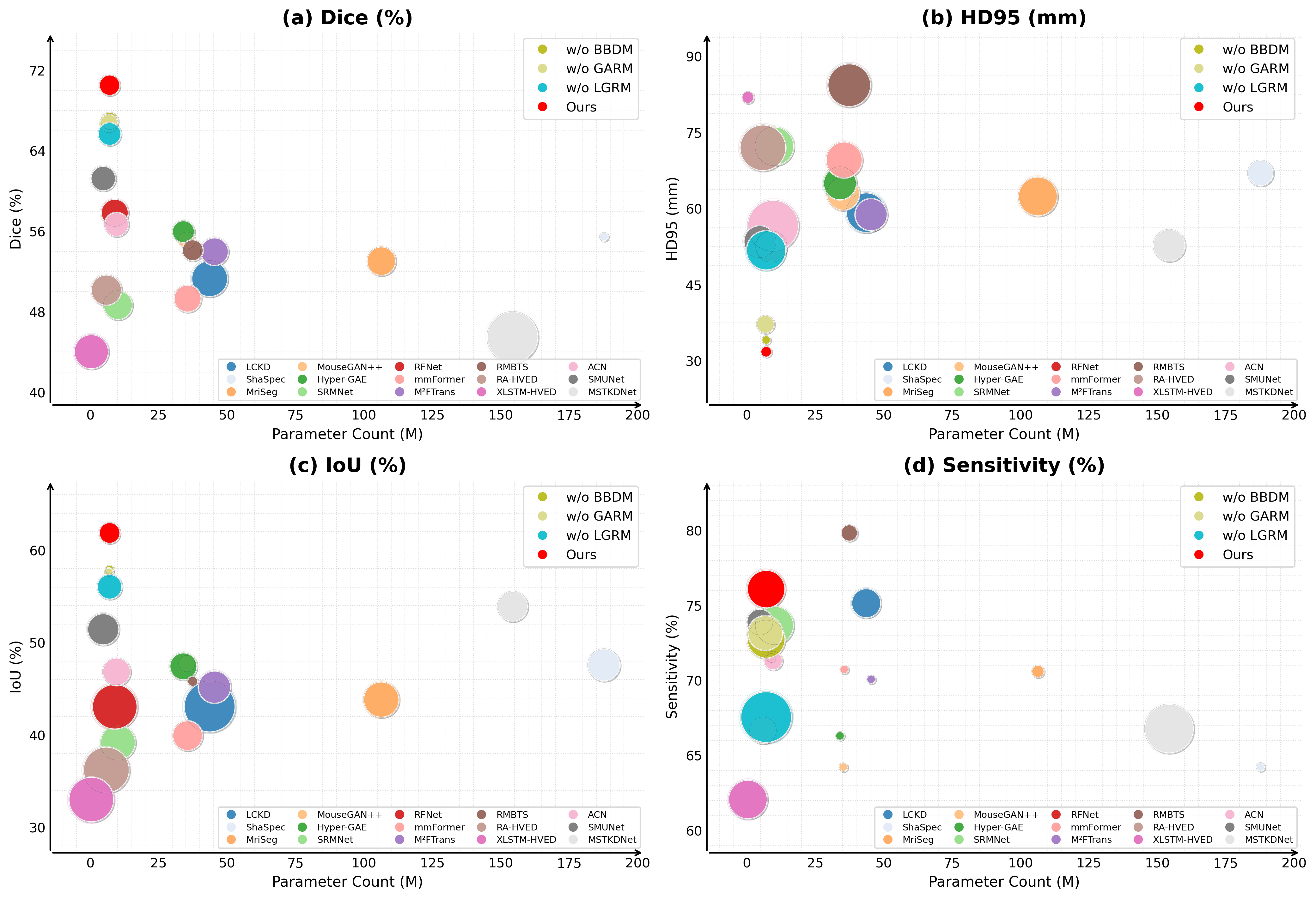}
  \caption{Scatter plots showing the relationship between model performance and parameter count across four evaluation metrics: (a) \textbf{Dice (\%)}, (b) \textbf{HD95 (mm)}, (c) \textbf{IoU (\%)}, and (d) \textbf{Sensitivity (\%)}. The proposed method (\textbf{Ours}) is highlighted in red.}
  \label{fig:params}
\end{figure}

\begin{figure}[!t]
  \centering
  \includegraphics[width=\textwidth]{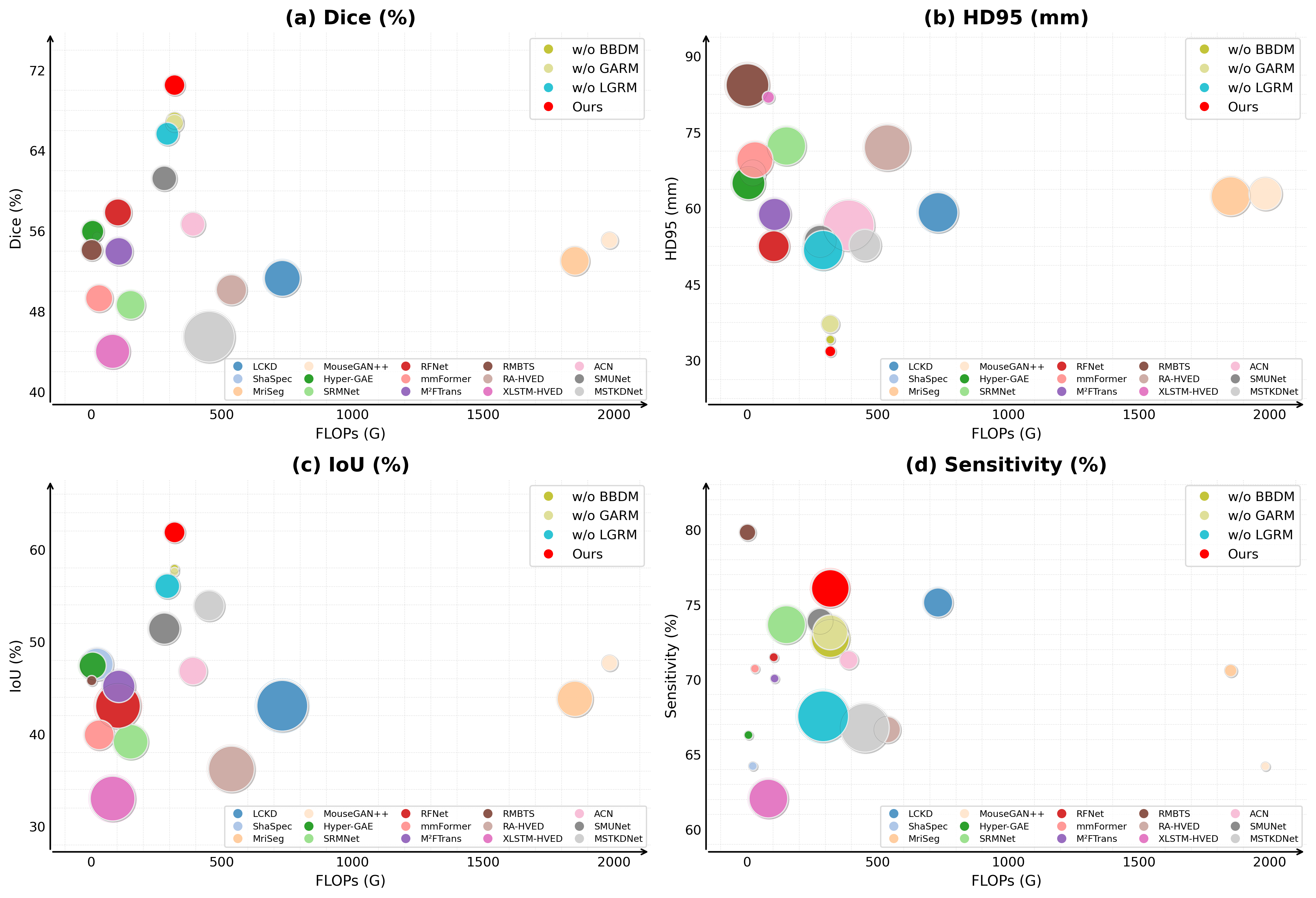}
  \caption{Scatter plots showing the relationship between model performance and FLOPs across four evaluation metrics: (a) \textbf{Dice (\%)}, (b) \textbf{HD95 (mm)}, (c) \textbf{IoU (\%)}, and (d) \textbf{Sensitivity (\%)}. The proposed method (\textbf{Ours}) is highlighted in red.}
  \label{fig:flops}
\end{figure}

\subsection{Assessment of Indicators}
To comprehensively evaluate the model’s segmentation performance on the BraTS 2024 and BraTS 2018 datasets, we adopted four commonly used metrics: Dice coefficient, Intersection over Union (IoU), Sensitivity, and Hausdorff Distance 95\% (HD95). Their definitions are as follows:

\begin{equation}
\begin{gathered}
\mathrm{Dice} = \frac{2\lvert X \cap Y \rvert}{\lvert X \rvert + \lvert Y \rvert}, \quad
\mathrm{IoU} = \frac{\lvert X \cap Y \rvert}{\lvert X \cup Y \rvert}, \quad
\mathrm{Sensitivity} = \frac{\mathrm{TP}}{\mathrm{TP} + \mathrm{FN}},\\[6pt]
d_{\mathrm H}(X,Y)=\max\{d_{XY},\, d_{YX}\}
          =\max\!\bigl(\max_{x\in X}\min_{y\in Y} d(x,y),\,
                       \max_{y\in Y}\min_{x\in X} d(x,y)\bigr),
\end{gathered}
\end{equation}
where $\lvert X \cap Y\rvert$ denotes the number of elements in the intersection of sets $X$ and $Y$, whereas $\lvert X\rvert$ and $\lvert Y\rvert$ represent the cardinalities of $X$ and $Y$, respectively; $x$ and $y$ are elements of $X$ and $Y$. The symbol $d$ denotes distance, and $\mathrm{HD95}$ is computed as the 95\textsuperscript{th} percentile of the distances between the boundary points of $X$ and $Y$. True Positives ($\mathrm{TP}$) and False Negatives ($\mathrm{FN}$) denote correctly predicted positive voxels and incorrectly predicted negative voxels.

\begin{figure}[htbp]
  \centering
  \includegraphics[width=\textwidth]{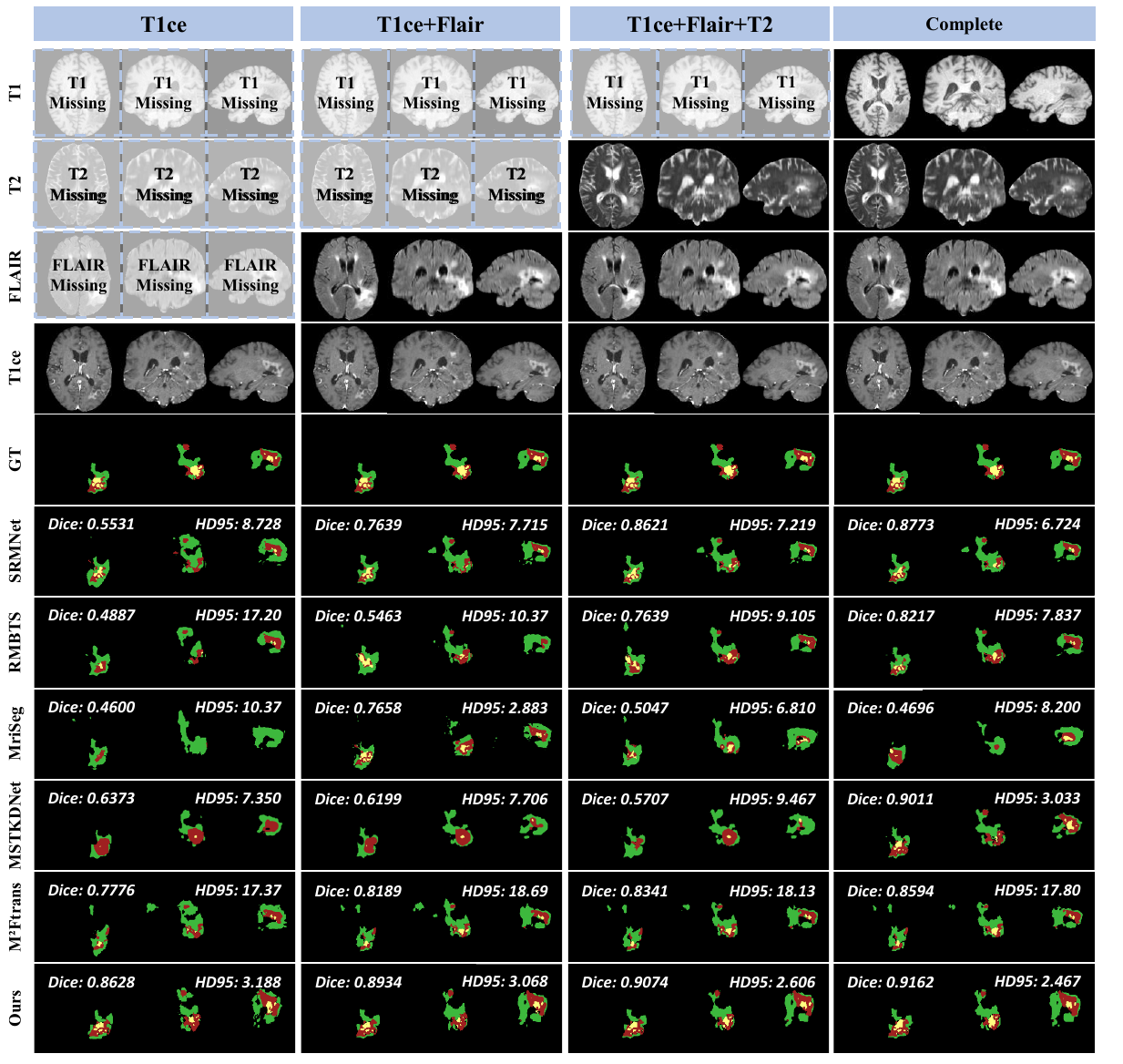}
\caption{\textbf{Qualitative comparison study on the BraTS 2024 dataset.} Visualization of a randomly selected sample from the BraTS 2024 dataset, showcasing segmentation results under different modality-missing combinations across three distinct axial views. The corresponding multimodal MRI sequences and corresponding ground truth are presented, along with outputs of models. Dice and HD95 metrics are displayed for each model-combination pair. From visualizations and metrics, our model (\textbf{Ours}) demonstrates superior segmentation accuracy and edge-control ability, with higher Dice values and lower HD95 results across various modality-missing scenarios. Color legend: WT = \textcolor[HTML]{9F2020}{red} + \textcolor{yellow}{yellow} + \textcolor[HTML]{3DB83D}{green}, TC = \textcolor[HTML]{9F2020}{red} + \textcolor{yellow}{yellow}, ET = \textcolor[HTML]{9F2020}{red}.}
  \label{fig:com}
\end{figure}
\subsection{Comparative Experiments}
We conducted systematic comparative experiments by randomly omitting input modalities to mimic clinical modality-missing scenarios, thus creating 15 input configurations encompassing single-, dual-, triple-, and full-modality combinations. As summarized in Tables~\ref{tab:dsc24}, \ref{tab:hd24}, \ref{tab:iou24}, \ref{tab:sen24},
\ref{tab:dsc18}, \ref{tab:hd18}, \ref{tab:iou18}, \ref{tab:sen18}, \ref{tab:dscpre}, \ref{tab:hdpre}, \ref{tab:ioupre}, \ref{tab:senpre}, we compared our method with several mainstream missing-modality brain-tumor segmentation approaches on the BraTS 2024 and BraTS 2018 datasets to verify the effectiveness of the proposed model. Owing to space constraints, this section focuses on the BraTS 2024 results; trends on BraTS 2018 are broadly similar. The results show that AdaMM consistently outperforms competing methods under most modality-missing settings, demonstrating superior robustness and segmentation accuracy.

Relative to data-generation methods such as MouseGAN++ \cite{c5yu2022mousegan++} and Hyper-GAE \cite{c6yang2023learning}, AdaMM raises the average single-modality Dice score by 23.55\% and 23.74\%, respectively, and improves the single-modality IoU score by 14.71\% and 17.37\%. These findings indicate that, instead of relying on complex generative networks to recover missing modalities, AdaMM leverages the BBDM distillation module to transfer structural and textural knowledge from the teacher model to the student model, achieving better performance without synthesizing additional data and with lower computational overhead.

Compared with feature-generation methods such as MriSeg \cite{c4ting2023multimodal} and ShaSpec \cite{c8wang2023multi}, AdaMM reduces the HD95 metric by 31.09mm and 35.7mm and increases Sensitivity by 5.94\% and 12.31\%. This result confirms that AdaMM’s GSME module aligns structural and style features between teacher and student more effectively, mitigating the information-reconstruction deficiencies caused by contextual absence in feature-generation approaches.

When contrasted with multi-task learning methods such as RA-HVED \cite{c23jeong2022region} and RMBTS \cite{c10chen2019robust}, we observe that these approaches often suffer feature conflicts within task coupling, limiting primary-task performance. By contrast, AdaMM’s LGRM module provides priors for the three tumor regions and effectively enhances target awareness. For example, under the FLAIR-only setting, AdaMM attains a Dice score of 59.91\%, exceeding the scores of RA-HVED and RMBTS 31.35\% and 37.56\%, respectively corresponding to improvements of 28.56\% and 22.35\%; the IoU metric rises by 28.88\% and 21.65\%, indicating superior reconstruction of tumor regions.

Against existing robustness-enhancement methods such as RFNet \cite{c11ding2021rfnet} and M$^{2}$FTrans \cite{c12shi2023mftrans}, AdaMM markedly improves feature representation under missing modalities via the GARM module’s graph-neural architecture. Unlike these methods, which directly process the raw incomplete inputs, AdaMM mitigates noise propagation and sample dependence by modeling modality combination relationships through a graph structure. It achieves gains of 76.54\% and 31.29 mm in average Sensitivity and HD95, respectively, underscoring its adaptability to detail recovery and boundary detection.

Compared with existing knowledge distillation methods such as ACN and SMUNet, AdaMM more effectively alleviates modality distribution inconsistency under missing-modality conditions, achieving simultaneous improvements in segmentation accuracy and boundary quality. SMUNet mitigates distribution shift via content-style disentanglement and style matching, but it is limited to feature-level alignment and lacks explicit modeling of cross-modality structural relationships, leading to unstable predictions for complex morphologies. In contrast, AdaMM explicitly propagates cross-modality topological relationships through GARM, while GSME further enforces global style alignment to preserve structural consistency. Compared with SMUNet, the Dice scores for WT, TC, and ET increase by 3.69\%, 13.42\%, and 11.10\%, respectively, while the corresponding HD95 values are reduced by 2.96 mm, 30.39 mm, and 33.29 mm. These gains are particularly pronounced in the TC and ET regions, highlighting the effectiveness of explicit structural modeling. 

Transformer-based compensation methods such as MMCFormer and MST-KDNet mainly rely on global self-attention for contextual alignment between full- and missing-modality inputs. However, AdaMM explicitly extracts cross-modality structural and topological knowledge through graph-guided reasoning, enabling more reliable alignment across diverse modality combinations. Compared with MMCFormer, AdaMM achieves improvements of 15.52\% and 4.09\% in IoU and Sensitivity, respectively, while reducing HD95 by 30.52 mm. These gains indicate the superiority of structured graph reasoning for boundary delineation under missing-modality conditions. In contrast to the implicit and unconstrained information propagation of global self-attention, which may amplify noisy modality cues, AdaMM constrains cross-modality information exchange via semantically consistent graph connections, effectively suppressing unreliable feature diffusion.

As shown in Fig. \ref{fig:params} and Fig. \ref{fig:flops}, AdaMM not only achieves superior segmentation performance across all tumor subregions, but also demonstrates the highest computational efficiency among the compared methods. Despite the introduction of graph-guided refinement and multi-level knowledge distillation mechanisms, the computational and parameter overhead of AdaMM remains low, requiring only 319.20 G FLOPs and 7.10M parameters. In addition to FLOPs and parameter counts, AdaMM processes a single 3D volume in 82 ms on a single NVIDIA Tesla A800 GPU, demonstrating practical efficiency for deployment. This favorable balance between accuracy and efficiency is primarily attributed to the lightweight design of the proposed modules. Specifically, GARM performs reasoning on a compact node-level graph representation rather than directly operating on high-dimensional, dense voxel features, thereby substantially reducing computational complexity. The adapter bank introduces only shallow residual branches to model modality-combination-specific feature distributions, and its parameter overhead is negligible relative to the backbone network. Meanwhile, the GSME and adversarial distillation losses are used exclusively during training for feature alignment and semantic guidance, and therefore incur no additional inference cost. In addition, LGRM consists of an extremely lightweight auxiliary branch whose computational burden is similarly minimal. As a result, AdaMM achieves competitive and even state-of-the-art segmentation performance without sacrificing computational efficiency, thereby validating its engineering feasibility and practical value for real-world deployment.


\begin{table*}[htbp]
  \centering  
  \caption{Quantitative comparison results measured in DSC (\%) on BraTS 2024}
  \label{tab:dsc24}

  \newcommand{\filledcirc}{\scalebox{1.6}{$\bullet$}}
  \newcommand{\opencirc}{\scalebox{1.6}{$\circ$}}
  \renewcommand{\arraystretch}{1.15}
  \setlength{\tabcolsep}{4pt}

\resizebox{\textwidth}{!}{%
\begin{tabular}{c|c|*{15}{c}|c|c}
\hline
\multirow{4}{*}{\textbf{Type}} & Flair & \opencirc&\opencirc&\opencirc&\filledcirc&\opencirc&\opencirc&\filledcirc&\opencirc&\filledcirc&\filledcirc&\filledcirc&\filledcirc&\filledcirc&\opencirc&\filledcirc & \multirow{4}{*}{\textbf{Avg.}} & \multirow{4}{*}{\textbf{Std.}}\\
                              & T1    & \opencirc&\opencirc&\filledcirc&\opencirc&\opencirc&\filledcirc&\filledcirc&\filledcirc&\opencirc&\opencirc&\filledcirc&\filledcirc&\opencirc&\filledcirc&\filledcirc & &\\
                              & T1c   & \opencirc&\filledcirc&\opencirc&\opencirc&\filledcirc&\filledcirc&\opencirc&\opencirc&\opencirc&\filledcirc&\filledcirc&\opencirc&\filledcirc&\filledcirc&\filledcirc & &\\
                              & T2    & \filledcirc&\opencirc&\opencirc&\opencirc&\filledcirc&\opencirc&\opencirc&\filledcirc&\filledcirc&\opencirc&\opencirc&\filledcirc&\filledcirc&\filledcirc&\filledcirc & &\\
\hline
\multirow{19}{*}{\textbf{WT}}  & LCKD \cite{c7wang2023learnable}       & 78.99 & \textbf{79.96} & \underline{}{75.62} & 77.98 & 78.68 & \underline{78.45} & 79.74 & 79.97 & 80.78 & 81.43 & 82.03 & 82.55 & 82.95 & \underline{82.91} & 83.28 & 80.15 & 2.29\\
                              & ShaSpec \cite{c8wang2023multi}      & 67.13 & 49.16 & 41.58 & 58.18 & 72.89 & 72.17 & 74.82 & 52.13 & 64.11 & 61.14 & 74.29 & 76.57 & 76.03 & 64.01 & 76.67 & 65.39 & 10.7\\
                              & MriSeg \cite{c4ting2023multimodal}  & 78.82 & 55.90 & 64.28 & 64.81 & \underline{81.81} & 70.84 & 72.80 & \underline{81.44} & 82.16 & 66.84 & 73.93 & 83.04 & 83.46 & \textbf{83.38} & 83.74 & 75.15 & 8.6\\
                              & MouseGAN++ \cite{c5yu2022mousegan++}       & 64.38 & 66.95 & 57.79 & 61.45 & 75.35 & 61.22 & 70.58 & 78.98 & 77.67 & 66.43 & 79.47 & 63.63 & 65.49 & 68.77 & 80.77 & 69.26 & 7.22\\
\hline
\end{tabular}}
\end{table*}

\begin{table*}[htbp]
  \centering   
  \caption{Quantitative comparison results measured in HD95 (mm) on BraTS 2024.}
  \label{tab:hd24}

  \newcommand{\filledcirc}{\scalebox{1.6}{$\bullet$}}
  \newcommand{\opencirc}{\scalebox{1.6}{$\circ$}}
  \renewcommand{\arraystretch}{1.15}
  \setlength{\tabcolsep}{4pt}

\resizebox{\textwidth}{!}{%
\begin{tabular}{c|c|*{15}{c}|c|c}
\hline
\multirow{4}{*}{\textbf{Type}} & Flair & \opencirc&\opencirc&\opencirc&\filledcirc&\opencirc&\opencirc&\filledcirc&\opencirc&\filledcirc&\filledcirc&\filledcirc&\filledcirc&\filledcirc&\opencirc&\filledcirc & \multirow{4}{*}{\textbf{Avg.}} & \multirow{4}{*}{\textbf{Std.}}\\
                              & T1    & \opencirc&\opencirc&\filledcirc&\opencirc&\opencirc&\filledcirc&\filledcirc&\filledcirc&\opencirc&\opencirc&\filledcirc&\filledcirc&\opencirc&\filledcirc&\filledcirc & &\\
                              & T1c   & \opencirc&\filledcirc&\opencirc&\opencirc&\filledcirc&\filledcirc&\opencirc&\opencirc&\opencirc&\filledcirc&\filledcirc&\opencirc&\filledcirc&\filledcirc&\filledcirc & &\\
                              & T2    & \filledcirc&\opencirc&\opencirc&\opencirc&\filledcirc&\opencirc&\opencirc&\filledcirc&\filledcirc&\opencirc&\opencirc&\filledcirc&\filledcirc&\filledcirc&\filledcirc & &\\
\hline
\multirow{19}{*}{\textbf{WT}}  & LCKD \cite{c7wang2023learnable} & \textbf{7.07} & \textbf{6.89} & \textbf{7.28} & 6.89 & \underline{6.55} & \textbf{6.70} & 6.41 & \textbf{6.29} & 6.11 & \underline{5.88} & \underline{5.69} & 5.54 & 5.40 & \underline{5.43} & \underline{5.32} & \underline{6.23} & \textbf{0.64} \\
& ShaSpec \cite{c8wang2023multi} & 28.84 & 40.88 & 47.63 & 25.32 & 23.85 & 22.57 & 19.72 & 44.79 & 23.95 & 30.51 & 20.97 & 16.57 & 16.76 & 29.69 & 14.29 & 27.09 & 9.88 \\
& MriSeg \cite{c4ting2023multimodal} & 10.61 & 23.63 & 12.72 & 12.27 & 8.40 & 12.97 & 8.54 & 8.26 & 6.48 & 11.81 & 9.88 & 6.23 & 7.80 & 6.37 & 7.22 & 10.21 & 4.24 \\
& MouseGAN++ \cite{c5yu2022mousegan++} & 27.50 & 27.02 & 13.69 & 21.72 & 15.61 & 14.98 & 13.85 & 19.90 & 15.46 & 12.21 & 14.62 & 18.09 & 16.67 & 13.94 & 11.25 & 17.10 & 4.78 \\
\hline
\end{tabular}}
\end{table*}

\begin{table*}[htbp]
  \centering    
  \caption{Quantitative comparison results measured in IoU (\%) on BraTS 2024.}
  \label{tab:iou24}

  \newcommand{\filledcirc}{\scalebox{1.6}{$\bullet$}}
  \newcommand{\opencirc}{\scalebox{1.6}{$\circ$}}
  \renewcommand{\arraystretch}{1.15}
  \setlength{\tabcolsep}{4pt}

\resizebox{\textwidth}{!}{%
\begin{tabular}{c|c|*{15}{c}|c|c}
\hline
\multirow{4}{*}{\textbf{Type}} & Flair & \opencirc&\opencirc&\opencirc&\filledcirc&\opencirc&\opencirc&\filledcirc&\opencirc&\filledcirc&\filledcirc&\filledcirc&\filledcirc&\filledcirc&\opencirc&\filledcirc & \multirow{4}{*}{\textbf{Avg.}} & \multirow{4}{*}{\textbf{Std.}}\\
                              & T1    & \opencirc&\opencirc&\filledcirc&\opencirc&\opencirc&\filledcirc&\filledcirc&\filledcirc&\opencirc&\opencirc&\filledcirc&\filledcirc&\opencirc&\filledcirc&\filledcirc & &\\
                              & T1c   & \opencirc&\filledcirc&\opencirc&\opencirc&\filledcirc&\filledcirc&\opencirc&\opencirc&\opencirc&\filledcirc&\filledcirc&\opencirc&\filledcirc&\filledcirc&\filledcirc & &\\
                              & T2    & \filledcirc&\opencirc&\opencirc&\opencirc&\filledcirc&\opencirc&\opencirc&\filledcirc&\filledcirc&\opencirc&\opencirc&\filledcirc&\filledcirc&\filledcirc&\filledcirc & &\\
\hline
\multirow{19}{*}{\textbf{WT}}  & LCKD \cite{c7wang2023learnable} & \textbf{69.49} & \underline{67.03} & \textbf{65.38} & 68.36 & 69.20 & 68.87 & 70.56 & 70.80 & 71.87 & 72.73 & 73.52 & 74.22 & 74.76 & \textbf{74.68} & 75.17 & 71.11 & \textbf{2.94} \\
& ShaSpec \cite{c8wang2023multi} & 67.13 & 49.16 & 41.58 & 58.18 & \textbf{72.89} & \textbf{72.17} & 74.82 & 52.13 & 64.11 & 61.14 & 74.29 & 76.57 & 76.03 & 64.01 & 76.67 & 65.39 & 10.70 \\
& MriSeg \cite{c4ting2023multimodal} & \underline{67.66} & 41.34 & 50.11 & 50.79 & \underline{71.64} & 57.27 & 59.8 & \underline{71.04} & 72.08 & 52.93 & 61.15 & 73.11 & 73.88 & 73.48 & 74.29 & 63.37 & 10.43 \\
& MouseGAN++ \cite{c5yu2022mousegan++} & 54.28 & 51.98 & 48.5 & 57.09 & 67.37 & 51.35 & 61.96 & \textbf{71.46} & 70.8 & 56.91 & 72.17 & 56.22 & 56.67 & 60.22 & 73.54 & 60.7 & 8.07 \\
\hline
\end{tabular}}
\end{table*}

\begin{table*}[htbp]
  \centering    
  \caption{Quantitative comparison results measured in Sensitivity (\%) on BraTS 2024.}
  \label{tab:sen24}

  \newcommand{\filledcirc}{\scalebox{1.6}{$\bullet$}}
  \newcommand{\opencirc}{\scalebox{1.6}{$\circ$}}
  \renewcommand{\arraystretch}{1.15}
  \setlength{\tabcolsep}{4pt}

\resizebox{\textwidth}{!}{%
\begin{tabular}{c|c|*{15}{c}|c|c}
\hline
\multirow{4}{*}{\textbf{Type}} & Flair & \opencirc&\opencirc&\opencirc&\filledcirc&\opencirc&\opencirc&\filledcirc&\opencirc&\filledcirc&\filledcirc&\filledcirc&\filledcirc&\filledcirc&\opencirc&\filledcirc & \multirow{4}{*}{\textbf{Avg.}} & \multirow{4}{*}{\textbf{Std.}}\\
                              & T1    & \opencirc&\opencirc&\filledcirc&\opencirc&\opencirc&\filledcirc&\filledcirc&\filledcirc&\opencirc&\opencirc&\filledcirc&\filledcirc&\opencirc&\filledcirc&\filledcirc & &\\
                              & T1c   & \opencirc&\filledcirc&\opencirc&\opencirc&\filledcirc&\filledcirc&\opencirc&\opencirc&\opencirc&\filledcirc&\filledcirc&\opencirc&\filledcirc&\filledcirc&\filledcirc & &\\
                              & T2    & \filledcirc&\opencirc&\opencirc&\opencirc&\filledcirc&\opencirc&\opencirc&\filledcirc&\filledcirc&\opencirc&\opencirc&\filledcirc&\filledcirc&\filledcirc&\filledcirc & &\\
\hline
\multirow{19}{*}{\textbf{WT}}  & LCKD \cite{c7wang2023learnable}       & \underline{81.27} & \underline{78.69} & 77.94 & 80.76 & 81.36 & 80.88 & 82.23 & 82.45 & 83.38 & 84.01 & 84.51 & 85.05 & 85.47 & \textbf{85.39} & 85.74 & 82.61 & 2.38 \\
& ShaSpec \cite{c8wang2023multi}      & 60.16 & 54.78 & 59.9 & 54.96 & 58.21 & 55.78 & 60.61 & 47.87 & 63.99 & 67.11 & 71.06 & 73.1 & 76.71 & 73.27 & 76.88 & 63.63 & 8.66 \\
& MriSeg \cite{c4ting2023multimodal}  & 69.23 & 64.8 & 66.91 & 66.03 & 69.66 & 67.52 & 67.73 & 69.99 & 69.05 & 65.89 & 66.75 & 69.57 & 69.64 & 69.52 & 69.91 & 68.15 & 1.67 \\
& MouseGAN++ \cite{c5yu2022mousegan++}       & 65.5 & 65.21 & 62.63 & 62.37 & 67.58 & 64.67 & 66.52 & 67.53 & 68.12 & 65.68 & 68.08 & 64.51 & 64.2 & 65.34 & 68.48 & 65.76 & 1.87 \\
\hline
\end{tabular}}
\end{table*}

\begin{table*}[htbp]
  \centering    
  \caption{Quantitative comparison results measured in DSC (\%) on BraTS 2018.}
  \label{tab:dsc18}

  \newcommand{\filledcirc}{\scalebox{1.6}{$\bullet$}}
  \newcommand{\opencirc}{\scalebox{1.6}{$\circ$}}
  \renewcommand{\arraystretch}{1.15}
  \setlength{\tabcolsep}{4pt}

\resizebox{\textwidth}{!}{%
\begin{tabular}{c|c|*{15}{c}|c|c}
\hline
\multirow{4}{*}{\textbf{Type}} & Flair & \opencirc&\opencirc&\opencirc&\filledcirc&\opencirc&\opencirc&\filledcirc&\opencirc&\filledcirc&\filledcirc&\filledcirc&\filledcirc&\filledcirc&\opencirc&\filledcirc & \multirow{4}{*}{\textbf{Avg.}} & \multirow{4}{*}{\textbf{Std.}}\\
                              & T1    & \opencirc&\opencirc&\filledcirc&\opencirc&\opencirc&\filledcirc&\filledcirc&\filledcirc&\opencirc&\opencirc&\filledcirc&\filledcirc&\opencirc&\filledcirc&\filledcirc & &\\
                              & T1c   & \opencirc&\filledcirc&\opencirc&\opencirc&\filledcirc&\filledcirc&\opencirc&\opencirc&\opencirc&\filledcirc&\filledcirc&\opencirc&\filledcirc&\filledcirc&\filledcirc & &\\
                              & T2    & \filledcirc&\opencirc&\opencirc&\opencirc&\filledcirc&\opencirc&\opencirc&\filledcirc&\filledcirc&\opencirc&\opencirc&\filledcirc&\filledcirc&\filledcirc&\filledcirc & &\\
\hline
\multirow{19}{*}{\textbf{WT}}&LCKD \cite{c7wang2023learnable}        & 76.18                & 71.32                & 70.72                & 74.31                & 75.28                & 74.93                & 76.73                & 77.26                & 78.31                & 79.16                & 79.94                & 80.59                & 81.08                & 81.12                & 81.56                & 77.23                & \textbf{3.35} \\
&ShaSpec \cite{c8wang2023multi}         & 78.89                & 42.74                & 43.99                & 66.55                & 81.97                & 82.15                & 82.34                & 56.70                & 69.91                & 71.98                & 83.19                & 83.53                & 83.75                & 73.32                & 84.41                & 72.36                & 13.70               \\
\hline
\end{tabular}}
\end{table*}

\begin{table*}[htbp]
  \centering    
  \caption{Quantitative comparison results measured in HD95 (mm) on BraTS 2018.}
  \label{tab:hd18}

  \newcommand{\filledcirc}{\scalebox{1.6}{$\bullet$}}
  \newcommand{\opencirc}{\scalebox{1.6}{$\circ$}}
  \renewcommand{\arraystretch}{1.15}
  \setlength{\tabcolsep}{4pt}

\resizebox{\textwidth}{!}{%
\begin{tabular}{c|c|*{15}{c}|c|c}
\hline
\multirow{4}{*}{\textbf{Type}} & Flair & \opencirc&\opencirc&\opencirc&\filledcirc&\opencirc&\opencirc&\filledcirc&\opencirc&\filledcirc&\filledcirc&\filledcirc&\filledcirc&\filledcirc&\opencirc&\filledcirc & \multirow{4}{*}{\textbf{Avg.}} & \multirow{4}{*}{\textbf{Std.}}\\
                              & T1    & \opencirc&\opencirc&\filledcirc&\opencirc&\opencirc&\filledcirc&\filledcirc&\filledcirc&\opencirc&\opencirc&\filledcirc&\filledcirc&\opencirc&\filledcirc&\filledcirc & &\\
                              & T1c   & \opencirc&\filledcirc&\opencirc&\opencirc&\filledcirc&\filledcirc&\opencirc&\opencirc&\opencirc&\filledcirc&\filledcirc&\opencirc&\filledcirc&\filledcirc&\filledcirc & &\\
                              & T2    & \filledcirc&\opencirc&\opencirc&\opencirc&\filledcirc&\opencirc&\opencirc&\filledcirc&\filledcirc&\opencirc&\opencirc&\filledcirc&\filledcirc&\filledcirc&\filledcirc & &\\
\hline
\multirow{19}{*}{\textbf{WT}} &LCKD \cite{c7wang2023learnable}        & 9.42                 & 10.62                & 9.46                 & 8.10                 & 7.69                 & \underline{7.34}        & 6.74                 & 6.67                 & 6.32                 & 6.01                 & 5.72                 & 5.52                 & 5.36                 & \underline{5.31}        & 5.16                 & 7.03                 & \underline{1.66}       \\
&ShaSpec \cite{c8wang2023multi}         & 30.29                & 67.62                & 74.98                & 35.17                & 24.01                & 17.89                & 24.43                & 52.28                & 26.95                & 23.77                & 11.55                & 21.43                & 13.67                & 16.91                & 10.21                & 30.08                & 19.09               \\
\hline
\end{tabular}}
\end{table*}

\begin{table*}[htbp]
  \centering   
  \caption{Quantitative comparison results measured in IoU (\%) on BraTS 2018.}
  \label{tab:iou18}

  \newcommand{\filledcirc}{\scalebox{1.6}{$\bullet$}}
  \newcommand{\opencirc}{\scalebox{1.6}{$\circ$}}
  \renewcommand{\arraystretch}{1.15}
  \setlength{\tabcolsep}{4pt}

\resizebox{\textwidth}{!}{%
\begin{tabular}{c|c|*{15}{c}|c|c}
\hline
\multirow{4}{*}{\textbf{Type}} & Flair & \opencirc&\opencirc&\opencirc&\filledcirc&\opencirc&\opencirc&\filledcirc&\opencirc&\filledcirc&\filledcirc&\filledcirc&\filledcirc&\filledcirc&\opencirc&\filledcirc & \multirow{4}{*}{\textbf{Avg.}} & \multirow{4}{*}{\textbf{Std.}}\\
                              & T1    & \opencirc&\opencirc&\filledcirc&\opencirc&\opencirc&\filledcirc&\filledcirc&\filledcirc&\opencirc&\opencirc&\filledcirc&\filledcirc&\opencirc&\filledcirc&\filledcirc & &\\
                              & T1c   & \opencirc&\filledcirc&\opencirc&\opencirc&\filledcirc&\filledcirc&\opencirc&\opencirc&\opencirc&\filledcirc&\filledcirc&\opencirc&\filledcirc&\filledcirc&\filledcirc & &\\
                              & T2    & \filledcirc&\opencirc&\opencirc&\opencirc&\filledcirc&\opencirc&\opencirc&\filledcirc&\filledcirc&\opencirc&\opencirc&\filledcirc&\filledcirc&\filledcirc&\filledcirc & &\\
\hline
\multirow{19}{*}{\textbf{WT}} &LCKD \cite{c7wang2023learnable}        & 66.33                & 61.06                & 60.35                & 64.55                & 65.70                & 65.20                & 67.38                & 67.88                & 69.15                & 70.19                & 71.17                & 71.96                & 72.55                & 72.56                & 73.09                & 67.94                & \underline{3.95}       \\
&ShaSpec \cite{c8wang2023multi}         & 66.93                & 29.31                & 30.29                & 52.96                & 70.62                & 70.82                & 71.34                & 41.69                & 56.37                & 58.10                & 72.11                & 72.82                & 73.10                & 59.76                & 73.94                & 60.01                & 14.85               \\
\hline
\end{tabular}}
\end{table*}

\begin{table*}[htbp]
  \centering    
  \caption{Quantitative comparison results measured in Sensitivity (\%) on BraTS 2018.}
  \label{tab:sen18}

  \newcommand{\filledcirc}{\scalebox{1.6}{$\bullet$}}
  \newcommand{\opencirc}{\scalebox{1.6}{$\circ$}}
  \renewcommand{\arraystretch}{1.15}
  \setlength{\tabcolsep}{4pt}

\resizebox{\textwidth}{!}{%
\begin{tabular}{c|c|*{15}{c}|c|c}
\hline
\multirow{4}{*}{\textbf{Type}} & Flair & \opencirc&\opencirc&\opencirc&\filledcirc&\opencirc&\opencirc&\filledcirc&\opencirc&\filledcirc&\filledcirc&\filledcirc&\filledcirc&\filledcirc&\opencirc&\filledcirc & \multirow{4}{*}{\textbf{Avg.}} & \multirow{4}{*}{\textbf{Std.}}\\
                              & T1    & \opencirc&\opencirc&\filledcirc&\opencirc&\opencirc&\filledcirc&\filledcirc&\filledcirc&\opencirc&\opencirc&\filledcirc&\filledcirc&\opencirc&\filledcirc&\filledcirc & &\\
                              & T1c   & \opencirc&\filledcirc&\opencirc&\opencirc&\filledcirc&\filledcirc&\opencirc&\opencirc&\opencirc&\filledcirc&\filledcirc&\opencirc&\filledcirc&\filledcirc&\filledcirc & &\\
                              & T2    & \filledcirc&\opencirc&\opencirc&\opencirc&\filledcirc&\opencirc&\opencirc&\filledcirc&\filledcirc&\opencirc&\opencirc&\filledcirc&\filledcirc&\filledcirc&\filledcirc & &\\
\hline
\multirow{19}{*}{\textbf{WT}} &LCKD \cite{c7wang2023learnable}        & 75.22                & 71.97                & 71.48                & 75.23                & 75.85                & 75.52                & 77.34                & 77.59                & 78.65                & 79.60                & 80.42                & 81.03                & 81.49                & 81.40                & 81.76                & 77.64                & 3.26                \\
&ShaSpec \cite{c8wang2023multi}         & 74.92                & 46.81                & 57.55                & 68.16                & 76.31                & 77.60                & 74.86                & 63.17                & 67.09                & 72.71                & 76.99                & 74.39                & 76.38                & 71.86                & 75.48                & 70.29                & 8.35                \\
\hline
\end{tabular}}
\end{table*}

\begin{table*}[htbp]
  \centering   
  \caption{Quantitative comparison results measured in DSC (\%) on Pretreat-MetsToBrain-Masks.}
  \label{tab:dscpre}

  \newcommand{\filledcirc}{\scalebox{1.6}{$\bullet$}}
  \newcommand{\opencirc}{\scalebox{1.6}{$\circ$}}
  \renewcommand{\arraystretch}{1.15}
  \setlength{\tabcolsep}{4pt}

\resizebox{\textwidth}{!}{%
\begin{tabular}{c|c|*{15}{c}|c|c}
\hline
\multirow{4}{*}{\textbf{Type}} & Flair & \opencirc&\opencirc&\opencirc&\filledcirc&\opencirc&\opencirc&\filledcirc&\opencirc&\filledcirc&\filledcirc&\filledcirc&\filledcirc&\filledcirc&\opencirc&\filledcirc & \multirow{4}{*}{\textbf{Avg.}} & \multirow{4}{*}{\textbf{Std.}}\\
                              & T1    & \opencirc&\opencirc&\filledcirc&\opencirc&\opencirc&\filledcirc&\filledcirc&\filledcirc&\opencirc&\opencirc&\filledcirc&\filledcirc&\opencirc&\filledcirc&\filledcirc & &\\
                              & T1c   & \opencirc&\filledcirc&\opencirc&\opencirc&\filledcirc&\filledcirc&\opencirc&\opencirc&\opencirc&\filledcirc&\filledcirc&\opencirc&\filledcirc&\filledcirc&\filledcirc & &\\
                              & T2    & \filledcirc&\opencirc&\opencirc&\opencirc&\filledcirc&\opencirc&\opencirc&\filledcirc&\filledcirc&\opencirc&\opencirc&\filledcirc&\filledcirc&\filledcirc&\filledcirc & &\\
\hline
\multirow{19}{*}{\textbf{WT}} & LCKD \cite{c7wang2023learnable}                    & 44.03                & 37.36                & 41.03                & 45.09                & 45.09                & 45.86                & 48.71                & 50.05                & 51.02                & 51.62                & 52.89                & 54.14                & 54.40                & 54.76                & 55.64                & 48.78                & 5.31                 \\
& ShaSpec \cite{c8wang2023multi}                     & 44.76                & 45.60                & 35.76                & 56.53                & 61.15                & 54.25                & 57.17                & 48.54                & 57.26                & 65.07                & 65.89                & 57.10                & 65.44                & 62.00                & 65.02                & 56.10                & 8.64                 \\
\hline
 \end{tabular}}
\end{table*}

\begin{table*}[htbp]
  \centering

  \caption{Quantitative comparison results measured in HD95 (mm) on Pretreat-MetsToBrain-Masks.}
  \label{tab:hdpre}

  \newcommand{\filledcirc}{\scalebox{1.6}{$\bullet$}}
  \newcommand{\opencirc}{\scalebox{1.6}{$\circ$}}
  \renewcommand{\arraystretch}{1.15}
  \setlength{\tabcolsep}{4pt}

\resizebox{\textwidth}{!}{%
\begin{tabular}{c|c|*{15}{c}|c|c}
\hline
\multirow{4}{*}{\textbf{Type}} & Flair & \opencirc&\opencirc&\opencirc&\filledcirc&\opencirc&\opencirc&\filledcirc&\opencirc&\filledcirc&\filledcirc&\filledcirc&\filledcirc&\filledcirc&\opencirc&\filledcirc & \multirow{4}{*}{\textbf{Avg.}} & \multirow{4}{*}{\textbf{Std.}}\\
                              & T1    & \opencirc&\opencirc&\filledcirc&\opencirc&\opencirc&\filledcirc&\filledcirc&\filledcirc&\opencirc&\opencirc&\filledcirc&\filledcirc&\opencirc&\filledcirc&\filledcirc & &\\
                              & T1c   & \opencirc&\filledcirc&\opencirc&\opencirc&\filledcirc&\filledcirc&\opencirc&\opencirc&\opencirc&\filledcirc&\filledcirc&\opencirc&\filledcirc&\filledcirc&\filledcirc & &\\
                              & T2    & \filledcirc&\opencirc&\opencirc&\opencirc&\filledcirc&\opencirc&\opencirc&\filledcirc&\filledcirc&\opencirc&\opencirc&\filledcirc&\filledcirc&\filledcirc&\filledcirc & &\\
\hline
\multirow{19}{*}{\textbf{WT}}  & LCKD \cite{c7wang2023learnable}                    & 30.99                & 37.37                & 37.41                & 35.80                & 34.77                & 32.57                & 30.67                & 28.75                & \underline{ 28.49}          & 28.53                & 27.42                & 26.48                & 26.26                & 25.28                & 24.63                & 30.36                & 4.18                \\
& ShaSpec \cite{c8wang2023multi}                     & 101.92               & 56.00                & 62.95                & 66.89                & 47.85                & 48.80                & 48.57                & 112.86               & 86.80                & 31.01                & 30.03                & 70.74                & 22.97                & 48.84                & 32.56                & 57.92                & 25.54               \\
\hline
\end{tabular}}
\end{table*}

\begin{table*}[htbp]
  \centering  
  \caption{Quantitative comparison results measured in IoU (\%) on Pretreat-MetsToBrain-Masks.}
  \label{tab:ioupre}

  \newcommand{\filledcirc}{\scalebox{1.6}{$\bullet$}}
  \newcommand{\opencirc}{\scalebox{1.6}{$\circ$}}
  \renewcommand{\arraystretch}{1.15}
  \setlength{\tabcolsep}{4pt}

\resizebox{\textwidth}{!}{%
\begin{tabular}{c|c|*{15}{c}|c|c}
\hline
\multirow{4}{*}{\textbf{Type}} & Flair & \opencirc&\opencirc&\opencirc&\filledcirc&\opencirc&\opencirc&\filledcirc&\opencirc&\filledcirc&\filledcirc&\filledcirc&\filledcirc&\filledcirc&\opencirc&\filledcirc & \multirow{4}{*}{\textbf{Avg.}} & \multirow{4}{*}{\textbf{Std.}}\\
                              & T1    & \opencirc&\opencirc&\filledcirc&\opencirc&\opencirc&\filledcirc&\filledcirc&\filledcirc&\opencirc&\opencirc&\filledcirc&\filledcirc&\opencirc&\filledcirc&\filledcirc & &\\
                              & T1c   & \opencirc&\filledcirc&\opencirc&\opencirc&\filledcirc&\filledcirc&\opencirc&\opencirc&\opencirc&\filledcirc&\filledcirc&\opencirc&\filledcirc&\filledcirc&\filledcirc & &\\
                              & T2    & \filledcirc&\opencirc&\opencirc&\opencirc&\filledcirc&\opencirc&\opencirc&\filledcirc&\filledcirc&\opencirc&\opencirc&\filledcirc&\filledcirc&\filledcirc&\filledcirc & &\\
\hline
\multirow{19}{*}{\textbf{WT}}  & LCKD \cite{c7wang2023learnable}                    & 36.53                & 29.83                & 31.96                & 36.30                & 36.54                & 36.59                & 39.37                & 40.40                & \underline{ 41.52}          & 42.34                & 43.56                & 44.74                & 45.11                & 45.29                & 46.10                & 39.75                & 4.84                 \\
& ShaSpec \cite{c8wang2023multi}                     & 36.90                & 34.21                & 26.54                & 47.66                & 50.01                & 41.97                & 48.31                & 39.97                & 48.57                & 54.35                & 55.41                & 48.53                & 54.92                & 51.01                & 54.63                & 46.20                & 8.27                 \\
\hline
\end{tabular}}
\end{table*}

\begin{table*}[htbp]
  \centering 

  \caption{Quantitative comparison results measured in SEN (\%) on Pretreat-MetsToBrain-Masks.}
  \label{tab:senpre}

  \newcommand{\filledcirc}{\scalebox{1.6}{$\bullet$}}
  \newcommand{\opencirc}{\scalebox{1.6}{$\circ$}}
  \renewcommand{\arraystretch}{1.15}
  \setlength{\tabcolsep}{4pt}

\resizebox{\textwidth}{!}{%
\begin{tabular}{c|c|*{15}{c}|c|c}
\hline
\multirow{4}{*}{\textbf{Type}} & Flair & \opencirc&\opencirc&\opencirc&\filledcirc&\opencirc&\opencirc&\filledcirc&\opencirc&\filledcirc&\filledcirc&\filledcirc&\filledcirc&\filledcirc&\opencirc&\filledcirc & \multirow{4}{*}{\textbf{Avg.}} & \multirow{4}{*}{\textbf{Std.}}\\
                              & T1    & \opencirc&\opencirc&\filledcirc&\opencirc&\opencirc&\filledcirc&\filledcirc&\filledcirc&\opencirc&\opencirc&\filledcirc&\filledcirc&\opencirc&\filledcirc&\filledcirc & &\\
                              & T1c   & \opencirc&\filledcirc&\opencirc&\opencirc&\filledcirc&\filledcirc&\opencirc&\opencirc&\opencirc&\filledcirc&\filledcirc&\opencirc&\filledcirc&\filledcirc&\filledcirc & &\\
                              & T2    & \filledcirc&\opencirc&\opencirc&\opencirc&\filledcirc&\opencirc&\opencirc&\filledcirc&\filledcirc&\opencirc&\opencirc&\filledcirc&\filledcirc&\filledcirc&\filledcirc & &\\
\hline
\multirow{19}{*}{\textbf{WT}}  & LCKD \cite{c7wang2023learnable}                    & 45.35                & 36.71                & 40.63                & 46.18                & 45.83                & 45.98                & 48.89                & 49.84                & \underline{ 51.08}          & 52.00                & 53.28                & 54.53                & 54.79                & 54.90                & 55.73                & 49.05                & 5.40                 \\
& ShaSpec \cite{c8wang2023multi}                     & 49.72                & 66.87                & 46.48                & 61.91                & 68.29                & 68.12                & 64.40                & 53.75                & 62.92                & 73.03                & 73.39                & 63.23                & 73.28                & 69.25                & 72.98                & 64.51                & 8.27                 \\
\hline
\end{tabular}}
\end{table*}

\subsection{Ablation Experiments}
\begin{figure}[htbp]
  \centering
  \includegraphics[width=\textwidth]{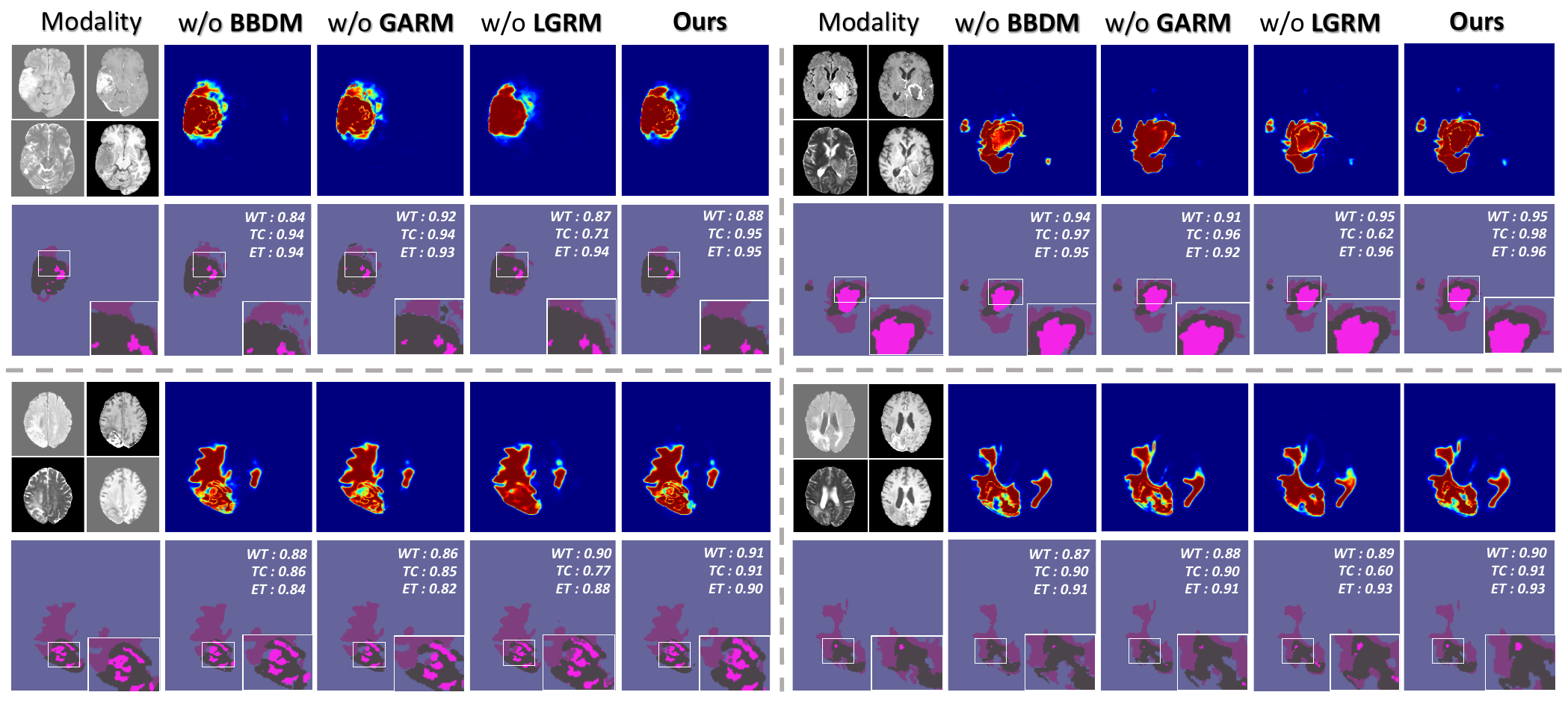}
    \caption{\textbf{Qualitative ablation study on the BraTS 2024 dataset.}
    Input images and ground-truth masks are shown for four modality configurations
    (T1, T1+T1Gd+T2+FLAIR, T1Gd+T2, and T1+T1Gd+T2).
    From left to right, the remaining columns present the segmentation outputs and
    corresponding prediction heatmaps produced by models without BBDM, without GARM,
    without LGRM, and with the complete model.
    Dice scores for WT, TC, and ET are reported.}
  \label{fig:abl}
\end{figure}

To verify the contribution of each module to overall performance, we conducted ablation experiments on the BraTS 2018 and BraTS 2024 datasets, successively removing BBDM, GARM, and LGRM for comparative analysis. Owing to space constraints, this section focuses on the results of BraTS 2024 dataset; trends on BraTS 2018 dataset are broadly similar. As shown in Tables~\ref{tab:dsc24}, \ref{tab:hd24}, \ref{tab:iou24}, \ref{tab:sen24}, 
\ref{tab:dsc18}, \ref{tab:hd18}, \ref{tab:iou18}, \ref{tab:sen18}, \ref{tab:dscpre}, \ref{tab:hdpre}, \ref{tab:ioupre}, \ref{tab:senpre}, \ref{tab:sensabla} and Fig. \ref{fig:abl} eliminating any single component leads to substantial declines in Dice, HD95, Sensitivity, and IoU, demonstrating that all three modules are essential for maintaining accuracy and robustness, and underscoring that their synergistic interaction is critical for ensuring model stability and generalization.

\begin{figure}[t]
  \centering
  \includegraphics[width=\textwidth]{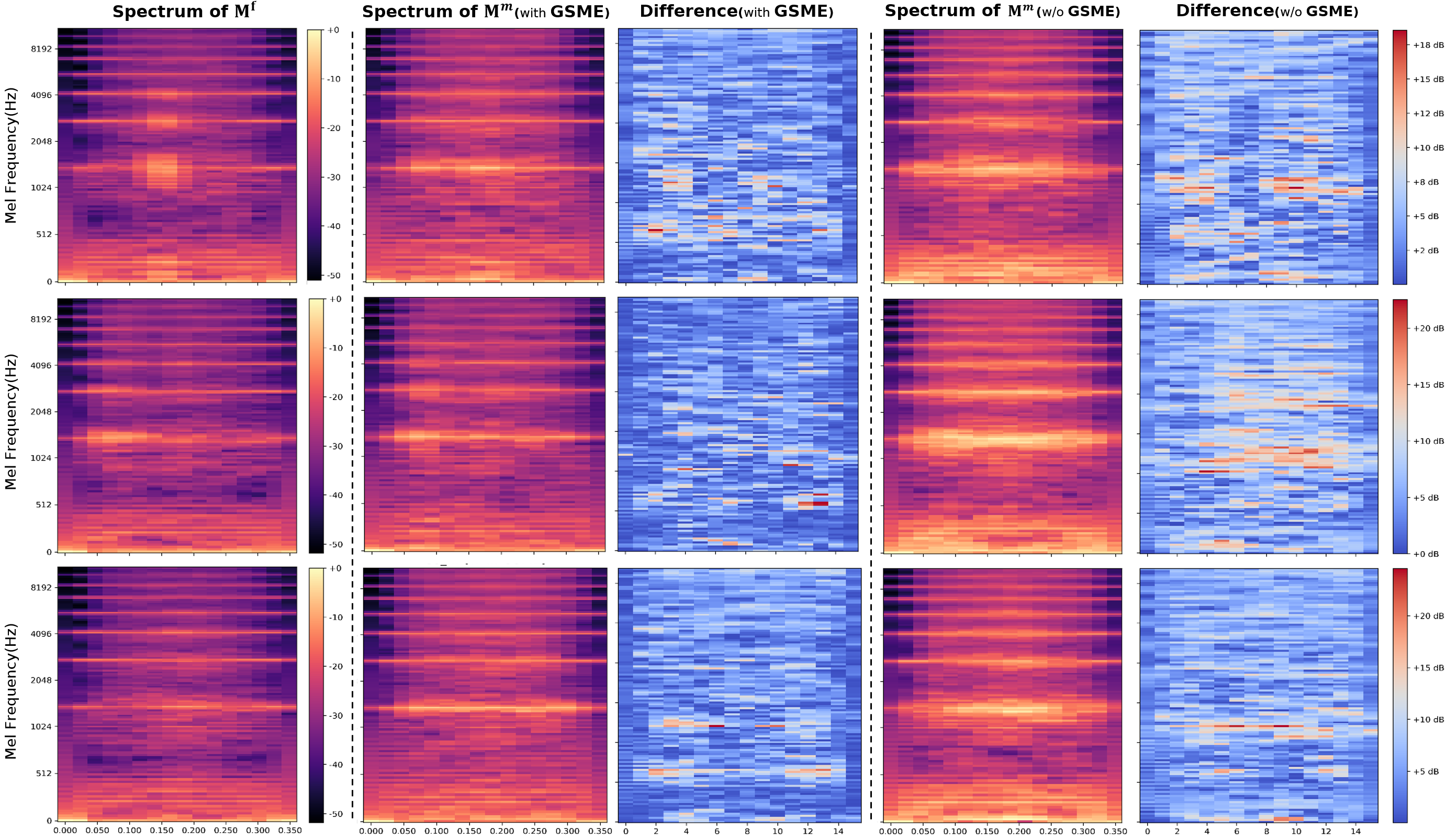}
  
\caption{\textbf{Comparison of Mel spectrograms between full-modality input and missing-modality reconstructions with and without GSME.} Three modality combinations are shown across rows: (1) T1C, (2) T1C+FLAIR, and (3) T1C+FLAIR+T1. The x-axis denotes time for spectrograms and index (0-14) for difference maps, while the y-axis indicates Mel frequency. }

  \label{fig:spec}
\end{figure}

\begin{table*}[htbp]
\centering
\fontsize{8pt}{10pt}\selectfont
\caption{Ablation Study on Loss Components and sensitivity analysis of the threshold $\tau$ across three datasets. The table compares segmentation performance under different $\tau$ values ($0.3$, $0.5$, $0.7$, $0.9$, and the default $\tau=0.8$) on BraTS 2024, BraTS 2018, and Pretreat-MetsToBrain-Masks, allowing a comprehensive assessment of how varying the threshold influences Dice, HD95, and Sensitivity.}

\label{tab:sensabla}
\begin{adjustbox}{width=1\textwidth, center}

\tiny
\renewcommand{\arraystretch}{0.9}
\begin{tabularx}{\textwidth}{
c   
*{3}{>{\centering\arraybackslash}X}
*{3}{>{\centering\arraybackslash}X}
*{3}{>{\centering\arraybackslash}X}
*{3}{>{\centering\arraybackslash}X}
}
\toprule
\multirow[c]{2}{*}{\textbf{Method}}

& \multicolumn{3}{c}{\textbf{Dice (\%$\uparrow$)}}
& \multicolumn{3}{c}{\textbf{HD95 (mm$\downarrow$)}}
& \multicolumn{3}{c}{\textbf{Sensitivity (\%$\uparrow$)}}
& \multicolumn{3}{c}{\textbf{IoU (\%$\uparrow$)}} \\
\cline{2-13}
\phantom{\textbf{Method}}
 & WT & TC & ET
 & WT & TC & ET
 & WT & TC & ET
 & WT & TC & ET \\
\hline

\rowcolor{red!15}    
\multicolumn{13}{c}{\textbf{\textit{\fontfamily{ptm}\selectfont BraTS 2024 dataset}}} \\
w/o \textbf{GSME}  
& 80.35 & 62.24 & 61.09 & 7.28 & \underline{45.64} & \textbf{42.86} & 82.91 & \underline{70.30} & \underline{77.03} & 54.21 & 54.00 & 51.90  \\
\rowcolor{gray!15}
w/o \textbf{Discriminator}  
& 79.63 & 63.00 & 62.11 & 7.14 & 45.64 & 43.10 & 82.21 & 69.50 & \textbf{78.33} & 51.42 & 53.69 & 50.86  \\ 

$\tau =\textbf{0.3}$      
& 77.21 & 59.32 & 59.09 & 10.77 & 53.20 & 55.78 & 77.53 & 66.51 & 64.39 & 67.99 & 52.24 & 52.20  \\
\rowcolor{gray!15}
$\tau = \textbf{0.5}$ 
& 80.66 & 62.64 & 59.32 & 7.21 & 51.17 & 49.42 & 80.92 & 67.49 & 66.11 & 59.00 & 53.45 & 52.49  \\
$\tau = \textbf{0.7}$  
& \underline{81.39} & \textbf{64.70} & 61.05 & \textbf{5.23} & 48.90 & 44.27 & \underline{83.80} & 69.28 & 57.84 & 84.28 & \textbf{56.93} & 53.01  \\
\rowcolor{gray!15}
$\tau = \textbf{0.9}$  
& 81.70 & 62.38 & \underline{62.32} & 6.92 & 46.23 & 45.12 & 82.11 & 66.32 & \underline{72.12} & \underline{71.95} & \underline{55.71} & \underline{54.45}  \\

Ours $(\tau = \textbf{0.8})$
& \textbf{83.90} & \underline{64.49} & \textbf{63.54}
& \underline{5.62} & \textbf{45.32} & \underline{42.94}
& \textbf{86.16} & \textbf{72.68} & 70.79
& \textbf{73.54} & 56.55 & \textbf{55.60}  \\

\rowcolor{blue!15}
\multicolumn{13}{c}{\textbf{\textit{\fontfamily{ptm}\selectfont BraTS 2018 dataset}}} \\

w/o \textbf{GSME}  
& 84.58 & 78.31 & 62.89 & 7.27 & \textbf{6.21} & 45.95  & \underline{89.03} & 68.67 & 53.29 & \underline{75.92} & 69.19 & 50.63  \\
\rowcolor{gray!15}
w/o \textbf{Discriminator}  
& 84.45 & 78.22 & 62.00 & 7.56 & 6.74 & 46.47 & 88.32 & 66.07 & 52.12 & 73.29 & 68.76 & 49.58  \\ 

$\tau = \textbf{0.3}$      
& 80.31 & 74.32 & 59.90 & 10.87 & 10.67 & 49.91 & 79.26 & 80.12 & 65.92 & 72.73 & 66.12 & 47.79 \\
\rowcolor{gray!15}
$\tau = \textbf{0.5}$ 
& 83.59 & 77.21 & 62.48 & 8.01 & 8.75 & 46.12 & 82.30 & \underline{82.54} & 68.31 & 74.92 & 66.24 & 50.41 \\
$\tau = \textbf{0.7}$  
& \underline{85.11} & 78.03 & \textbf{64.10} & 7.94 & 6.82 & \underline{42.22} & 81.82 & 82.31 & \textbf{70.12} & 75.87 & 68.73 & \textbf{51.88} \\
\rowcolor{gray!15}
$\tau = \textbf{0.9}$  
& 83.92 & \underline{78.92} & 61.87 & 8.28 & 9.20 & 45.05 & 82.12 & 81.40 & 66.43 & 75.72 & \underline{69.41} & 49.43 \\
Ours $(\tau = \textbf{0.8})$
& \textbf{86.94} & \textbf{80.81} & \underline{63.75} & \textbf{6.32} & \underline{6.46} & \textbf{38.04} & \textbf{89.26} & \textbf{83.60} & \underline{69.47} & \textbf{77.03} & \textbf{70.60} & \underline{51.31} \\

\rowcolor{yellow!15}
\multicolumn{13}{c}{\textbf{\textit{\fontfamily{ptm}\selectfont Pretreat-MetsToBrain-Masks dataset}}} \\

w/o \textbf{GSME}  
& 62.33 & 62.36 & 53.11 & 34.76 & 40.10 & 41.22 & \underline{68.11} & \underline{68.91} & 57.55 & 52.68 & \underline{52.77} & \underline{40.35}  \\
\rowcolor{gray!15}
w/o \textbf{Discriminator}  
& 62.53 & \underline{62.77} & \underline{53.37} & \underline{31.00} & 37.29 & 38.34 & 68.26 & 69.30 & \underline{58.02} & \underline{53.21} & 52.98 & 40.84  \\ 

$\tau = \textbf{0.3}$      
& 61.23 & 58.34 & 48.94 & 37.00 & 49.02 & 46.23 & 60.21 & 55.95 & 51.09 &  44.12 & 41.18 & 37.40 \\
\rowcolor{gray!15}
$\tau = \textbf{0.5}$ 
& 63.26 & 61.73 & 49.35 & 35.19 & 41.77 & 37.98 & 61.60 & 58.07 & 52.12 & 46.26 & 44.64 & 36.76 \\
$\tau = \textbf{0.7}$  
& \underline{64.99} & 62.43 & 50.04 & 33.69 & 40.91 & 37.92 & 63.22 & 57.31 & 51.30 &  48.14 & 45.38 & 38.37 \\
\rowcolor{gray!15}
$\tau = \textbf{0.9}$  
& 62.83 & 58.90 & 47.11 & 34.85 & \underline{38.30} & \underline{35.23} & 62.98 & 65.00 & 51.22 &  45.80 & 41.74 & 35.81 \\
Ours $(\tau = \textbf{0.8})$
& \textbf{65.34} & \textbf{63.14} & \textbf{53.50} 
& \textbf{25.12} & \textbf{27.98} & \textbf{27.57}
& \textbf{71.71} & \textbf{73.85}  & \textbf{59.72} 
& \textbf{52.71} & \textbf{53.01} & \textbf{41.25}  \\

\bottomrule
\end{tabularx}
\end{adjustbox}
\end{table*}

\subsubsection{Ablation of the Bi-Bottleneck Distillation Module}
When BBDM including the GSME feature-distillation branch and the teacher-student discriminator was removed, the Dice scores for ET, WT, and TC fell by 5.16\%, 3.01\%, and 2.82\%, respectively, while HD95 rose to 3.72 mm, 2.71 mm, and 2.05 mm; Sensitivity and IoU likewise declined markedly. Without GSME, the student model loses structural-textural alignment guidance and cannot exploit the priors of complete modalities to compensate for missing information. Eliminating the discriminator removes style regularization, causing shallow features to drift over time and the decoder to miss fine textures, thereby leading to fragmented or jagged boundaries. The absence of adversarial regularization also weakens robustness to input noise and modality perturbations. These results underscore BBDM’s central role in facilitating knowledge transfer and maintaining resilience.

After removing the GSME loss, as reported in Table \ref{tab:sensabla}, the Dice scores for ET, WT, and TC decrease by 3.55\%, 2.25\%, and 2.45\%, respectively, while HD95 increases by 1.66 mm, 0.32 mm, and 0.08 mm, with Sensitivity and IoU exhibiting consistent degradation. In the absence of bottleneck-level alignment constraints on global style and structural characteristics between the teacher and student models, the student features undergo pronounced distributional drift under missing-modality conditions, making it difficult to stably reconstruct modality-related contrast and texture information.

\begin{figure}[t]
  \centering
  \includegraphics[width=\textwidth]{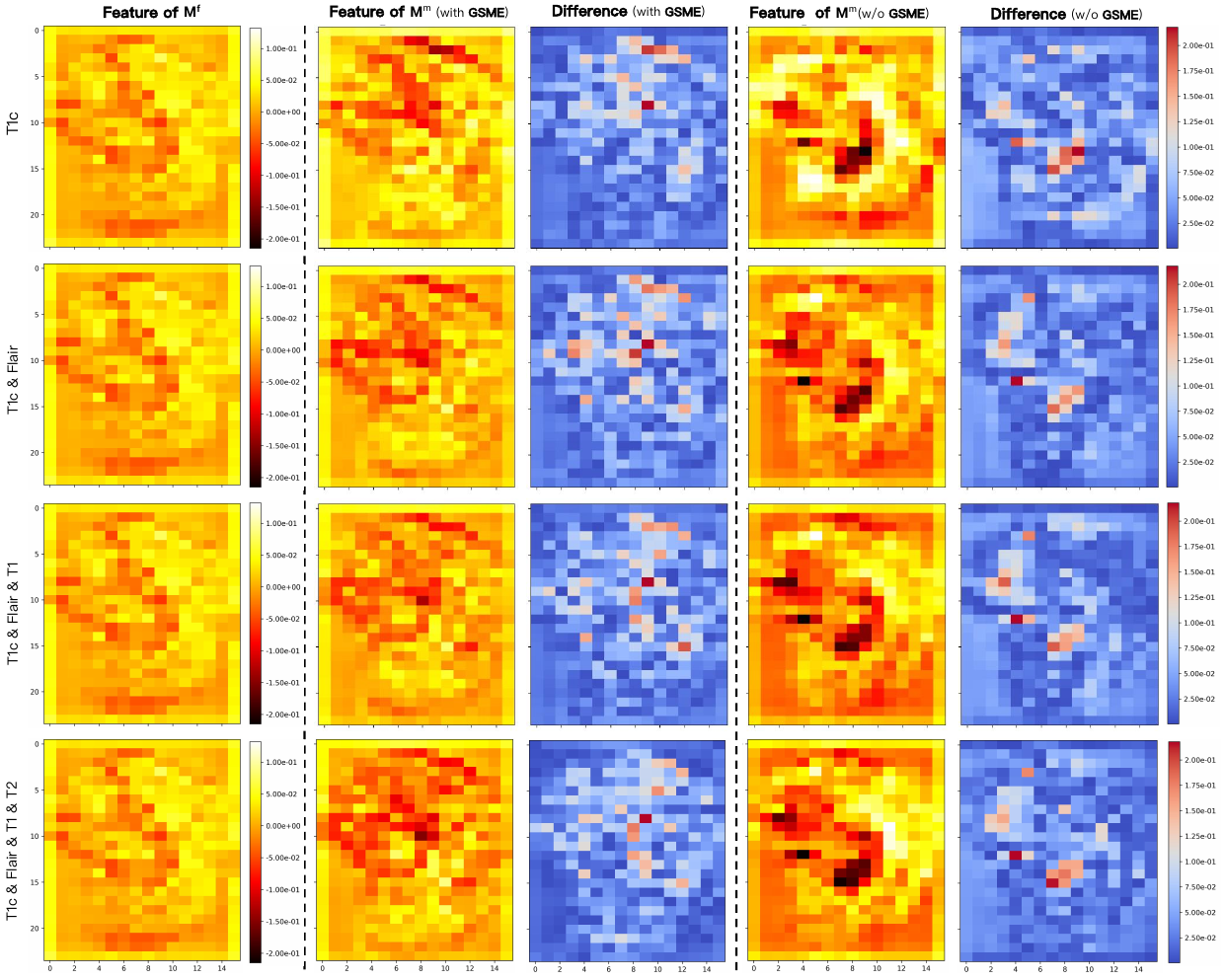}
  
\caption{\textbf{Comparison of features between full-modality input and missing-modality reconstructions with and without GSME.} Three modality combinations of a same sample are shown across rows: (1) T1C, (2) T1C+FLAIR, (3) T1C+FLAIR+T1, and (4) T1C+FLAIR+T1+T2.}

  \label{fig:feature}
\end{figure}

As shown in Fig. \ref{fig:spec} and Fig. \ref{fig:feature}, GSME is critical for stabilizing frequency- and spatial-domain representations. In the frequency domain, GSME maintains consistent low- to mid-frequency energy distributions between student and teacher features and suppresses disordered high-frequency components. Without GSME, mid- to high-frequency responses scatter, indicating instability in boundary- and texture-level representations. In the spatial domain, GSME encourages concentrated feature responses and stronger structural continuity with the teacher. Its absence leads to dispersed activations, jittery boundaries, and local spurious responses, which degrade structural delineation and boundary localization.

Removing the discriminator eliminates style regularization, causing shallow feature representations to drift over the course of training and leading the decoder to miss fine-grained texture details, which results in fragmented boundaries. The lack of adversarial regularization further reduces robustness to input noise and modality perturbations. As shown in Table \ref{tab:sensabla}, the Dice scores for WT, TC, and ET decrease by 4.27\%, 1.49\%, and 1.43\%.

\subsubsection{Ablation of the Graph-guided Adaptive Refinement Module}
Removing GARM reduced the Dice scores for ET, WT, and TC by 5.01\%, 3.53\%, and 3.25\%, with HD95 increasing to 3.00 mm, 7.71 mm, and 7.03 mm; Sensitivity and IoU dropped in parallel. Without the modality-combination residuals (CF) from the Adapter Bank and their graph-based alignment with general features (GF), the model’s ability to adapt to different missing-modality combinations weakens. Structural guidance is lost, breaking connectivity in internal regions (e.g., necrotic core, enhancing zones) and diminishing deformation recovery. During sliding-window refinement the coarse mask cannot adjust to modality-specific contrast and texture, causing edge collapse and more false positives, which harms both Sensitivity and IoU. These findings confirm GARM as a key module for structural recovery and cross-combination robustness.

To further investigate the influence of the similarity threshold $\tau$, we conducted a sensitivity analysis by setting $\tau$ to 0.6, 0.7, 0.8, and 0.9 across three datasets and multiple evaluation metrics. As shown in Table \ref{tab:sensabla}, $\tau = 0.8$ achieves the best performance across datasets and evaluation criteria. When $\tau$ is set to a lower value, such as 0.6 or 0.7, the resulting cross-graph adjacency becomes overly dense, introducing a large number of low-confidence inter-graph connections. This leads to the propagation of semantically weak or noisy information, which interferes with attention-based aggregation and degrades the learning of discriminative features. In contrast, setting $\tau$ to a higher value, such as 0.9, yields an excessively sparse cross-graph structure. This restricts effective information flow and limits the extent of cross-modality feature compensation. By balancing semantic reliability, $\tau = 0.8$ provides a effective operating point for cross-graph reasoning. This setting preserves high-confidence semantic connections while suppressing the propagation of unreliable information, thereby enabling stable feature refinement and consistent semantic alignment under missing-modality conditions.

\begin{table*}[t]
\centering
\tiny
\caption{Ablation Study on adapters and fusion method of features across three datasets. The table compares segmentation performance under different components on BraTS 2024, BraTS 2018, and Pretreat-MetsToBrain-Masks, allowing a comprehensive assessment of adapters and fusion methods influence Dice, HD95, and Sensitivity. }
\label{tab:ada}
\begin{adjustbox}{width=1\textwidth, center}

\renewcommand{\arraystretch}{0.9}
\begin{tabularx}{\textwidth}{
c   
*{3}{>{\centering\arraybackslash}X}
*{3}{>{\centering\arraybackslash}X}
*{3}{>{\centering\arraybackslash}X}
*{3}{>{\centering\arraybackslash}X}
}
\toprule
\multirow[c]{2}{*}{\textbf{Method}}
& \multicolumn{3}{c}{\textbf{Dice (\%$\uparrow$)}}
& \multicolumn{3}{c}{\textbf{HD95 (mm$\downarrow$)}}
& \multicolumn{3}{c}{\textbf{Sensitivity (\%$\uparrow$)}}
& \multicolumn{3}{c}{\textbf{IoU (\%$\uparrow$)}} \\
\cline{2-13}
\phantom{\textbf{Method}}
 & WT & TC & ET
 & WT & TC & ET
 & WT & TC & ET
 & WT & TC & ET \\
\hline

\rowcolor{red!15}    
\multicolumn{13}{c}{\textbf{\textit{\fontfamily{ptm}\selectfont BraTS 2024 dataset}}} \\

w/ \textbf{shared adapter}  
& 81.26 & 62.03 & 61.17 
& 5.97 & 46.88 & 44.72 
& 83.92 & 70.11 & 68.73 
& \underline{71.82} & \underline{54.98} & \underline{53.64}  \\

\rowcolor{gray!15}
w/ \textbf{linear fusion} 
& \underline{82.12} & \underline{62.81} & \underline{61.88} 
& \textbf{5.71} & \underline{46.97} & \underline{44.21} 
& \underline{85.47} & \underline{70.04} & \underline{69.36} 
& 71.02 & 53.21 & 53.18  \\ 

Ours
& \textbf{83.90} & \textbf{64.49} & \textbf{63.54}
& \underline{5.62} & \textbf{45.32} & \textbf{42.94}
& \textbf{86.16} & \textbf{72.68} & \textbf{70.79}
& \textbf{73.54} & \textbf{56.55} & \textbf{55.60} \\

\rowcolor{blue!15}
\multicolumn{13}{c}{\textbf{\textit{\fontfamily{ptm}\selectfont BraTS 2018 dataset}}} \\

w/ \textbf{shared adapter}  
& \underline{84.92} & 78.63 & 61.28 
& 7.18 & \underline{7.02} & 39.51 
& 87.41 & 81.72 & 66.83 
& \underline{74.36} & \underline{68.95} & \underline{49.27}  \\

\rowcolor{gray!15}
w/ \textbf{linear fusion} 
& 84.21 & \underline{79.34} & \underline{61.92} 
& \underline{6.74} & 7.38 & \underline{40.96} 
& \underline{88.73} & \underline{81.91} & \underline{67.12} 
& 73.42 & 68.08 & 49.03  \\ 

Ours
& \textbf{86.94} & \textbf{80.81} & \textbf{63.75} 
& \textbf{6.32} & \textbf{6.46} & \textbf{38.04} 
& \textbf{89.26} & \textbf{83.60} & \textbf{69.47} 
& \textbf{77.03} & \textbf{70.60} & \textbf{51.31} \\

\rowcolor{yellow!15}
\multicolumn{13}{c}{\textbf{\textit{\fontfamily{ptm}\selectfont Pretreat-MetsToBrain-Masks dataset}}} \\

w/ \textbf{shared adapter}  
& 61.92 & 59.86 & 49.73 
& 26.08 & 28.91 & 28.44 
& 68.47 & 70.92 & 56.84 
& 49.88 & 50.29 & 38.74  \\

\rowcolor{gray!15}
w/ \textbf{linear fusion} 
& \underline{63.71} & \underline{61.95} & \underline{52.04} 
& \underline{25.69} & \underline{28.54} & \underline{27.95} 
& \underline{70.26} & \underline{72.11} & \underline{58.21} 
& \underline{51.37} & \underline{52.03} & \underline{40.18}  \\ 

Ours
& \textbf{65.34} & \textbf{63.14} & \textbf{53.50} 
& \textbf{25.12} & \textbf{27.98} & \textbf{27.57}
& \textbf{71.71} & \textbf{73.85}  & \textbf{59.72} 
& \textbf{52.71} & \textbf{53.01} & \textbf{41.25}  \\
\bottomrule
\end{tabularx}
\end{adjustbox}
\end{table*}

To clarify the contribution of the modality-combination-aware adaptive mechanism, we have added two groups of minimally controlled ablation studies. In the first group, we replaced the original modality-combination-specific adapters with a single shared adapter. In the second group, we substituted the graph-guided refinement module with a linear, element-wise aggregation strategy. This change examines whether structured graph modeling is a necessary design component. The results are presented in Table \ref{tab:ada}. Empirical evidence shows that both simplified variants consistently and noticeably degrade performance. More specifically, a shared or coarsely grouped adapter fails to adequately capture the distributional discrepancies across diverse modality combinations, thereby diminishing the model’s capacity to perform targeted compensation for non-dominant modality configurations. Meanwhile, removing the graph structure, although reducing computational complexity through lightweight linear or element-wise fusion, impairs the model’s ability to effectively characterize the structural semantic relationships between combination-aware features and modality-invariant representations. Consequently, this results in inferior performance in boundary delineation and the discrimination of anatomically complex regions. Taken together, these findings substantiate that the performance gains of AdaMM predominantly stem from its explicitly designed combination-aware adaptive mechanism tailored to varying missing-modality patterns, rather than from a mere aggregation of auxiliary components. The corresponding experimental results and analyses have been incorporated into the revised manuscript for clarity and completeness.\\

\begin{table*}[t]
\centering
\tiny
\caption{Ablation Study on Loss Components and sensitivity analysis of the threshold $\boldsymbol{\lambda}$ across three datasets. The table compares segmentation performance under different $\boldsymbol{\lambda}$ values ($\boldsymbol{\lambda}=\lambda_1:\lambda_2:\lambda_3$) on BraTS 2024, BraTS 2018, and Pretreat-MetsToBrain-Masks. }

\label{tab:lambda}
\begin{adjustbox}{width=1\textwidth, center}

\renewcommand{\arraystretch}{0.9}
\begin{tabularx}{\textwidth}{
c   
*{3}{>{\centering\arraybackslash}X}
*{3}{>{\centering\arraybackslash}X}
*{3}{>{\centering\arraybackslash}X}
*{3}{>{\centering\arraybackslash}X}
}
\toprule
\multirow[c]{2}{*}{\textbf{Method}}

& \multicolumn{3}{c}{\textbf{Dice (\%$\uparrow$)}}
& \multicolumn{3}{c}{\textbf{HD95 (mm$\downarrow$)}}
& \multicolumn{3}{c}{\textbf{Sensitivity (\%$\uparrow$)}}
& \multicolumn{3}{c}{\textbf{IoU (\%$\uparrow$)}} \\
\cline{2-13}
\phantom{\textbf{Method}}
 & WT & TC & ET
 & WT & TC & ET
 & WT & TC & ET
 & WT & TC & ET \\
\hline

\rowcolor{red!15}    
\multicolumn{13}{c}{\textbf{\textit{\fontfamily{ptm}\selectfont BraTS 2024 dataset}}} \\

$\boldsymbol{\lambda} = [0.5, 1, 1]$
& 79.42 & 61.27 & 60.88 
& 9.96 & 52.34 & 54.12 
& 80.21 & 67.84 & 65.72 
& 70.88 & \underline{55.91} & \underline{54.77}  \\

\rowcolor{gray!15}
$\boldsymbol{\lambda} = [1, 0.5, 1]$
& \underline{82.76} & \textbf{65.11} & \underline{62.83} 
& \textbf{5.34} & \underline{47.68} & \underline{43.61} 
& \underline{84.91} & \underline{70.92} & \textbf{71.33} 
& \underline{72.91} & 55.02 & 54.44  \\

$\boldsymbol{\lambda} = [1, 1, 0.5]$
& 82.03 & 63.18 & 61.94 
& 6.88 & 49.83 & 47.76 
& 82.61 & 69.43 & 67.98 
& 72.74 & 54.08 & 54.21  \\

\rowcolor{gray!15}
Ours $(\boldsymbol{\lambda} = [1, 1, 1])$
& \textbf{83.90} & \underline{64.49} & \textbf{63.54}
& \underline{5.62} & \textbf{45.32} & \textbf{42.94}
& \textbf{86.16} & \textbf{72.68} & \underline{70.79}
& \textbf{73.54} & \textbf{56.55} & \textbf{55.60} \\

\rowcolor{blue!15}
\multicolumn{13}{c}{\textbf{\textit{\fontfamily{ptm}\selectfont BraTS 2018 dataset}}} \\

$\boldsymbol{\lambda} = [0.5, 1, 1]$
& 83.12 & 76.54 & 60.92 
& 8.14 & 8.01 & 40.83 
& 85.43 & 79.62 & 66.84 
& 73.66 & 67.12 & 48.96  \\

\rowcolor{gray!15}
$\boldsymbol{\lambda} = [1, 0.5, 1]$
& \underline{86.11} & \textbf{80.94} & \underline{63.48} 
& \textbf{6.28} & \textbf{6.33} & \underline{38.72} 
& \underline{88.41} & \textbf{83.92} & \textbf{70.12} 
& \underline{76.92} & \underline{69.88} & \underline{51.03}  \\

$\boldsymbol{\lambda} = [1, 1, 0.5]$
& 85.21 & 78.92 & 62.11 
& 7.03 & 7.28 & 39.64 
& 87.02 & 81.34 & 68.12 
& 75.88 & 69.84 & 50.21  \\

\rowcolor{gray!15}
Ours $(\boldsymbol{\lambda} = [1, 1, 1])$
& \textbf{86.94} & \underline{80.81} & \textbf{63.75} 
& \underline{6.32} & \underline{6.46} & \textbf{38.04} 
& \textbf{89.26} & \underline{83.60} & \underline{69.47} 
& \textbf{77.03} & \textbf{70.60} & \textbf{51.31} \\

\rowcolor{yellow!15}
\multicolumn{13}{c}{\textbf{\textit{\fontfamily{ptm}\selectfont Pretreat-MetsToBrain-Masks dataset}}} \\

$\boldsymbol{\lambda} = [0.5, 1, 1]$
& 62.11 & 58.74 & 50.01 
& 27.61 & 30.33 & 29.88 
& 68.21 & 71.02 & 56.83 
& 50.02 & 50.11 & 38.74  \\

\rowcolor{gray!15}
$\boldsymbol{\lambda} = [1, 0.5, 1]$
& \underline{64.91} & \underline{62.88} & \underline{52.94} 
& \underline{25.94} & \underline{28.62} & 29.11
& \underline{71.12} & \underline{73.21} & {57.08} 
& \underline{52.21} & \underline{52.77} & \underline{40.92}  \\

$\boldsymbol{\lambda} = [1, 1, 0.5]$
& 63.98 & 61.02 & 51.72 
& 26.74 & 29.47 & \underline{28.82} 
& 70.03 & 72.58 & \underline{58.21} 
& 51.63 & 52.02 & 40.03  \\

\rowcolor{gray!15}
Ours $(\boldsymbol{\lambda} = [1, 1, 1])$
& \textbf{65.34} & \textbf{63.14} & \textbf{53.50} 
& \textbf{25.12} & \textbf{27.98} & \textbf{27.57}
& \textbf{71.71} & \textbf{73.85}  & \textbf{59.72} 
& \textbf{52.71} & \textbf{53.01} & \textbf{41.25}  \\
\bottomrule
\end{tabularx}
\end{adjustbox}
\end{table*}

\subsubsection{Ablation of the Lesion-presence Guided Reliability Module}
Removing LGRM substantially lowered Dice for ET, WT, and TC by 3.77\%, 5.12\%, and 6.02\% and raised HD95 to 3.19 mm, 26.94 mm, and 31.29 mm; Sensitivity and IoU deteriorated accordingly. Without lesion-presence probability guidance, pronounced boundary shifts arise during ROI cropping and sliding-window stitching; low-contrast areas and vascular artifacts are frequently misclassified as lesions, thereby markedly increasing false positives. Uncertain regions lack BCE feedback, yielding blurred masks, missing ET details, and disjoint WT envelopes, which impair 3D visualization and volume estimation. Thus, LGRM not only boosts aggregate performance as an auxiliary branch but also supplies dynamic reliability constraints essential for accuracy and cross-domain robustness.

\subsubsection{Ablation of the weights for losses}
To evaluate the sensitivity of the proposed framework to loss weighting, we conducted additional fixed-weight experiments on the BraTS 2024 dataset using $\boldsymbol{\lambda}=[0.5,1,1]$, $[1,0.5,1]$, $[1,1,0.5]$, and $[1,1,1]$, where $\boldsymbol{\lambda}$ denotes the relative weights of the three loss terms in the form $\lambda_1:\lambda_2:\lambda_3$. As shown in Table \ref{tab:lambda}, the performance variations across different settings are limited, indicating stable behavior under different weighting strategies. The equal-weight setting $\boldsymbol{\lambda}=[1,1,1]$ achieves the best overall performance, yielding the highest Dice for WT and ET with values of 83.90\% and 63.54\% and the lowest HD95 for TC and ET with values of 45.32 mm and 42.94 mm. These results indicate that the loss terms are complementary and that the model does not rely on delicate manual tuning of loss weights.

\section{Conclusion}
This paper proposes AdaMM, a multi-modal brain tumor segmentation framework designed for missing-modality scenarios, addressing the robustness and generalization challenges caused by incomplete MRI inputs in clinical settings. Centered on knowledge distillation, AdaMM integrates three complementary modules: GARM explicitly models semantic associations between general and modality-specific features to improve adaptability; BBDM enables precise bottleneck feature transfer between teacher and student networks via style alignment and adversarial learning; and LGRM incorporates lesion-aware priors through an auxiliary classification task to suppress false positives under missing modalities. We construct a brain tumor segmentation benchmark for missing-modality scenarios using the Pretreat-MetsToBrain-Masks and BraTS 2018, 2024 datasets, and systematically compare six major categories of existing methods. Experimental results demonstrate that AdaMM consistently outperforms state-of-the-art approaches under both complete and incomplete modality settings, validating the effectiveness of its design and offering practical guidance for future research. 

This work contributes a robust and generalizable solution for multi-modal medical image segmentation under incomplete inputs.  Although the current instantiation is evaluated on BraTS-style brain tumor segmentation, it is important to emphasize that not all components are inherently task-specific. The proposed framework consists of both task-adaptive and generalizable modules. Future directions include integrating self-supervised learning, extending to other organs and tasks, and enabling efficient deployment in resource-limited or time-constrained clinical environments.

\section*{CRediT authorship contribution statement}

\textbf{Shenghao Zhu}: Methodology, Visualization, Writing - original draft.  
\textbf{Yifei Chen}: Conceptualization, Methodology, Writing - original draft.  
\textbf{Weihong Chen}: Validation, Visualization.  
\textbf{Shuo Jiang}: Validation, Visualization.  
\textbf{Guanyu Zhou}: Validation, Visualization.  
\textbf{Yuanhan Wang}: Validation, Visualization.  
\textbf{Feiwei Qin}: Supervision, Writing - review \& editing.  
\textbf{Changmiao Wang}: Writing - review \& editing.  
\textbf{Qiyuan Tian}: Project administration, Writing - review \& editing.

\section*{Declaration of competing interest}
The authors declare that they have no known competing financial interests or personal relationships that could have appeared to influence the work reported in this paper.

\section*{Acknowledgments}
This work was supported by the National Natural Science Foundation of China (No. 82302166), Tsinghua University Startup Fund, Fundamental Research Funds for the Provincial Universities of Zhejiang (No. GK259909299001-006), Anhui Provincial Joint Construction Key Laboratory of Intelligent Education Equipment and Technology (No. IEET202401), and Guangdong Basic and Applied Basic Research Foundation (No. 2025A1515011617).

\bibliographystyle{elsarticle-num-names}
\bibliography{ref}
\end{document}